%% file: main.tex
\documentclass[sigconf]{acmart}

\usepackage{times}
\usepackage{latexsym}

\usepackage[T1]{fontenc}

\usepackage[utf8]{inputenc}

\usepackage{microtype}

\usepackage{url}

\usepackage{graphicx}        
\usepackage{multicol}        
\usepackage[bottom]{footmisc}

\usepackage{newtxtext}       %
\usepackage{newtxmath}       
\usepackage{caption}
\usepackage{makecell}
\usepackage{graphics}
\usepackage{booktabs}
\usepackage{times}
\usepackage{multirow}
\usepackage{enumitem}
\usepackage{amsthm}
\usepackage{amsmath,bm}

\usepackage{amssymb}
\usepackage{pifont}
\usepackage{tablefootnote}
\usepackage{threeparttable}
\usepackage{color, colortbl}
\colorlet{shadecolor}{gray!20}
\usepackage{soul}
\usepackage{tabularx}
\usepackage{array}
\usepackage{subcaption}

\usepackage{xcolor}  
\usepackage{tikz}    

\newcommand{\coloredcheckmark}{%
  \begin{tikzpicture}[baseline=(check.base)]
    \node[draw, fill=green, circle, inner sep=2pt] (check) {\textcolor{white}{\ding{52}}};
  \end{tikzpicture}%
}

\newcommand{\coloredcross}{%
  \begin{tikzpicture}[baseline=(cross.base)]
    \node[draw, fill=red, circle, inner sep=2pt] (cross) {\textcolor{white}{\ding{55}}};
  \end{tikzpicture}%
}

\newcommand{\model}{Structure Guided Prompt}

\AtBeginDocument{%
  \providecommand\BibTeX{{%
    \normalfont B\kern-0.5em{\scshape i\kern-0.25em b}\kern-0.8em\TeX}}}

\setcopyright{acmlicensed}
\copyrightyear{2018}
\acmYear{2018}
\acmDOI{XXXXXXX.XXXXXXX}

\acmConference[Conference acronym 'XX]{Make sure to enter the correct
  conference title from your rights confirmation emai}{June 03--05,
  2018}{Woodstock, NY}
\acmISBN{978-1-4503-XXXX-X/18/06}

\begin{document}

\title{\model{}: Instructing Large Language Model in Multi-Step Reasoning by Exploring Graph Structure of the Text}

\author{Kewei Cheng}
\affiliation{%
  \institution{UCLA}
  \city{Los Angeles}
  \state{CA}
  \country{USA}
}
\email{viviancheng@cs.ucla.edu}

\author{Nesreen K. Ahmed}
\affiliation{%
  \institution{Intel Labs}
\city{Santa Clara}
  \state{CA}
  \country{USA}
}
\email{nesreen.k.ahmed@intel.com}

\author{Theodore Willke}
\affiliation{%
  \institution{Intel Labs}
    \city{Portland}
  \state{OR}
  \country{USA}
}
\email{ted.willke@intel.com}

\author{Yizhou Sun}
\affiliation{%
 \institution{UCLA}
  \city{Los Angeles}
  \state{CA}
  \country{USA}
 }
\email{yzsun@cs.ucla.edu }




\renewcommand{\shortauthors}{Cheng, et al.}

\input{content/abstract}





\maketitle

\input{content/intro}

\input{content/related_work}
\input{content/method}
\input{content/tasks}

\input{content/experiment}

\input{content/discussion}
\input{content/conclusion}

\clearpage
\bibliographystyle{ACM-Reference-Format}
\bibliography{ref}

\input{content/appendix}

\end{document}

%% file: content/abstract.tex
\begin{abstract}
Although Large Language Models (LLMs) excel at addressing straightforward reasoning tasks, they frequently struggle with difficulties when confronted by more complex multi-step reasoning due to a range of factors. 
Firstly, natural language often encompasses complex relationships among entities, making it challenging to maintain a clear reasoning chain over longer spans. Secondly, the abundance of linguistic diversity means that the same entities and relationships can be expressed using different terminologies and structures, complicating the task of identifying and establishing connections between multiple pieces of information. Graphs provide an effective solution to represent data rich in relational information and capture long-term dependencies among entities. To harness the potential of graphs, our paper introduces \textit{\model{}}, an  innovative three-stage \textbf{task-agnostic} prompting framework designed to improve the multi-step reasoning capabilities of LLMs in a \textbf{zero-shot setting}. This framework explicitly converts unstructured text into a graph via LLMs and instructs them to navigate this graph using task-specific strategies to formulate responses. By effectively organizing information and guiding navigation, it enables LLMs to provide more accurate and context-aware responses. Our experiments show that this framework significantly enhances the reasoning capabilities of LLMs, enabling them to excel in a broader spectrum of natural language scenarios.


\end{abstract}

%% file: content/intro.tex
\section{Introduction}
\begin{figure*}[ht]
    \centering
    \includegraphics[width= \linewidth]{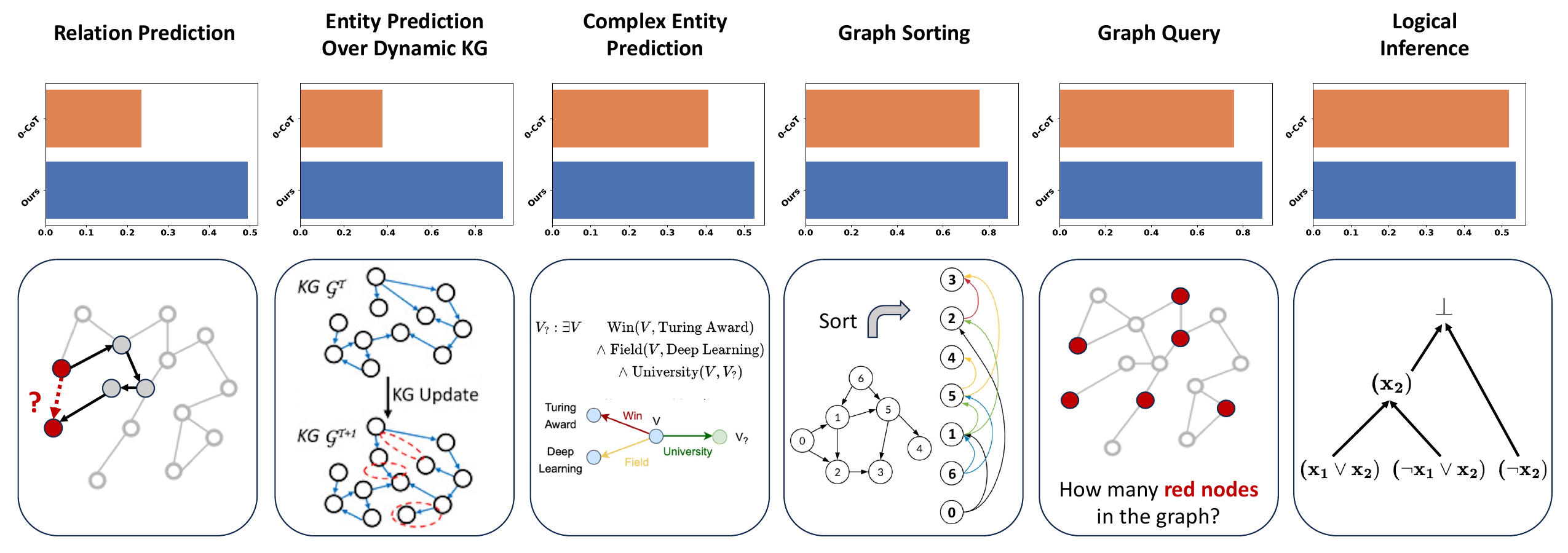}
    \caption{GPT-4’s performance using {\color{orange}0-shot chain-of-thought (0-CoT) (represented as the orange bars)} compared to the results of {\color{blue}\textit{\model{}} (represented as the blue bars)} across a variety of tasks. It is evident that \textit{\model{}} consistently and significantly outperforms the approach with 0-CoT.} 
    \label{fig:tasks}
\end{figure*}
Natural Language Processing (NLP) has witnessed significant advancements in recent years with the emergence of Large Language Models (LLMs) such as GPT-3~\cite{brown2020language} and ChatGPT~\cite{OpenAI2023GPT4TR}. These models have achieved remarkable results in tasks such as language generation, translation, and summarization~\cite{zhao2023survey}. However, studies have indicated that while LLMs can effectively handle straightforward reasoning problems, they often encounter challenges when faced with more complex reasoning, such as scenarios demanding multi-step reasoning~\cite{paranjape2023art}. 


Multi-step reasoning typically involves making inferences or answering questions that require multiple steps of logical reasoning. Here’s an illustration of multi-step reasoning: ``Marian went shoe shopping with her sister Michelle. Darnell's grandfather, Stanley, taught her how to make a paper airplane while her mother, Marian, prepared dinner. What is the family relationship between Michelle and Stanley?'' Various methods, such as chain-of-thought (CoT)~\cite{wei2022chain, saparov2022language} and Zero-Shot-CoT~\cite{kojima2022large}, have been proposed to improve multi-step reasoning in LLMs. These approaches involve step-by-step reasoning, either by providing examples with detailed intermediate steps leading to a conclusion or by prompting the model with ``Let’s think step by step'' in a zero-shot setting. Despite their effectiveness, LLMs still face challenges in effectively addressing complex multi-step reasoning questions.
The first challenge involves accurately comprehending relationships conveyed through natural language, as evident in the given example where Marian has a sister named Michelle, and Darnell has a grandfather named Stanley. Identifying these relationships accurately is crucial, but the inherent ambiguity in natural language makes this difficult. For instance, consider the sentence ``Darnell's grandfather, Stanley, taught her how to make a paper airplane while her mother, Marian, prepared dinner,'' correctly inferring that Marian, not Stanley, is Darnell's mother requires understanding the gender implications and the contextual relational information.
Second, LLMs must identify relevant information while ignoring the irrelevant. In the example ``Marian went shoe shopping with her sister Michelle,'' recognizing that Michelle is Marian's sister is crucial, while the detail about shoe shopping is not. This requires discernment in filtering out unnecessary details that could mislead.
Third, accurate multi-step reasoning requires LLMs to logically connect information. In the given scenario, two steps of inference are required. Initially, recognizing that Marian is Stanley's daughter, followed by combining this with the fact that Michelle is Marian's sister, leads to the final deduction that Michelle is Stanley's daughter. This process, typically more straightforward in formal logic due to clear logical indicators, becomes more complex in natural language due to the lack of explicit logical connectors, posing challenges for LLMs in constructing accurate reasoning paths.

\begin{figure}[h]
    \centering
    \includegraphics[width= \linewidth]{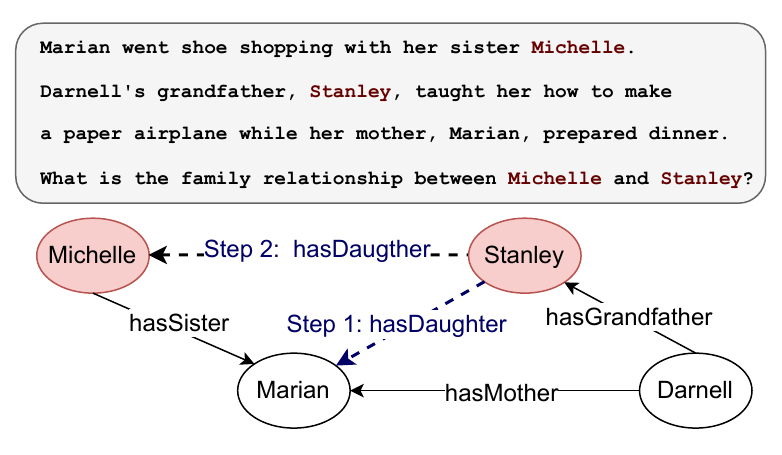}
    \caption{An example illustrating how humans manage multi-step questions. Our objective is to deduce the relationship between two individuals, Michelle and Stanley, highlighted in red, from a given story. Given that the story involves various individuals, humans typically first create a graph to clearly visualize the relationships among them. Then, they infer the relationship step by step, based on the graph.
    } 
    \label{fig:intro_example}
\end{figure}

Considering all the previously mentioned challenges, performing multi-step reasoning directly based on unstructured text is a challenging task. To reduce the complexity, \textbf{can LLMs be guided to adopt a more systematic and structured method for identifying reasoning paths for multi-step reasoning?}
Multi-step reasoning is essential to human intelligence, inspiring how we guide LLMs. Humans usually rely on structured knowledge representations, like Knowledge Graphs (KGs), to link different pieces of information clearly and systematically. Consider the question involving Marian, Michelle, Stanley, and Darnell.  Answering it can be challenging due to the multiple individuals involved and the need to remember and correctly sequence their relationships. To address this, humans often create a graph to visually represent the relationships, as depicted in Fig.~\ref{fig:intro_example}. They then deduce the relationships step by step, based on this graph. Although this method might seem simple, it is particularly effective, especially with longer inference chains.

Due to the advantages brought by KGs, there have been efforts to harness the strengths of both KGs and LLMs. These efforts typically involve integrating additional KGs as external tools to augment the reasoning capabilities of LLMs~\cite{pan2024unifying}. However, constructing and maintaining these KGs can be expensive, and using an external KG may overwhelm LLMs with too much irrelevant information
when addressing specific queries. In contrast, our approach takes a unique path. We firmly believe that natural language paragraphs inherently contain sufficient information for effectively answering questions. Rather than relying on external KGs, our approach centers on refining the organization of information within these paragraphs to enhance information comprehension and reasoning. Consequently, we introduce \textit{\model{}}, a novel prompting framework designed to guide LLMs in multi-step reasoning. It explicitly converts unstructured text into a graph and instructs LLMs to navigate this graph to formulate responses in a \textit{zero-shot setting}.
Acknowledging the diversity of queries and their corresponding graph structures, we have categorized reasoning tasks into various categories as shown in Fig.~\ref{fig:tasks}. Each category is aligned with a unique graph structure. These categories present distinctive challenges for LLMs. In the evaluation, we compared the performance of both GPT-3.5 and the advanced GPT-4 model~\cite{OpenAI2023GPT4TR} when equipped with our proposed prompt. Remarkably, our framework emerges as a catalyst, significantly enhancing the reasoning capabilities of LLMs across broader natural language scenarios. The results unequivocally demonstrate that \textit{\model{}} empowers general-purpose LLMs to achieve competitive performance, underscoring its pivotal role in exploring the graph structure of text for instructing LLMs in multi-step reasoning. In summary, our contribution can be categorized into three main aspects:
\begin{itemize}
    \item We propose \textit{\model{}}, a novel prompting framework designed to enhance the reasoning capability of LLMs by exploring the graph structure underlying the text. Within this framework, we delve into several distinct tasks, each tailored to specific graph structures.
    \item We show with experiments that our framework boosts the reasoning capability of general-purpose LLMs across a broader spectrum of natural language scenarios. 
    \item We conduct thorough analytical investigations, summarizing not only key open questions but also providing valuable insights for future research in this field. We hope these insights will inspire further exploration in the realm of reasoning.
\end{itemize}

%% file: content/related_work.tex
\section{Related Works}
\subsection{Multi-step Reasoning with LLMs}
Multi-step reasoning is a challenging NLP task that requires a system to make multiple inference steps to answer a question. 
While LLMs exhibit strong capabilities in one-hop inference, they struggle to perform effectively in multi-step reasoning. Numerous strategies have been suggested to enhance the multi-step reasoning capabilities of LLMs, such as implementing step-by-step reasoning using few-shot examples. Unlike ``naive'' prompting, which expects that the input should be immediately followed by the output or answer, eliciting prompts direct LLMs to tackle tasks by guiding them through intermediate steps before making predictions for the final output or answer. This method, known as chain-of-thought (CoT)~\cite{wei2022chain, saparov2022language}, has demonstrated that elicitive prompting equips LMs with superior reasoning abilities in a few-shot setting. Later, Zero-Shot-CoT~\cite{kojima2022large} presented similar capabilities in a zero-shot setting. They simply prepended the input question with the phrase ``Let’s think step by step'' before querying the model, and showed that large LMs performed well in zero-shot-CoT on reasoning tasks like GSM8K, though not as proficiently as in few-shot-CoT. Least to Most prompting (LtM)~\cite{zhou2022least} takes CoT prompting a step further by first breaking a problem into sub problems and then proceeds to solve each one independently. These sub-question answers are then synthesized to obtain the final response. Additionally, Tree of Thoughts (ToT)~\cite{yao2023tree} and Graph of Thoughts (GoT)~\cite{besta2023graph} use complex structures like trees and graphs to organize thoughts. These systems combine the way LLMs generate thoughts with search algorithms for systematic exploration, further enhancing their multi-step reasoning capabilities.
In contrast to all these approaches, our work introduces a distinct three-step prompting framework. This framework emulates the problem-solving approach employed by humans when dealing with data rich in relationships. It enables users to transform a natural language paragraph into a graph, and subsequently, based on the query type, navigate this graph for the purpose of answering questions.



\subsection{Integrate LLMs with Logical Inference}
The most studied approach to reasoning since the earliest days of AI is logical inference~\cite{carnap2012introduction}. Logical systems are fundamentally rule-based~\cite{quinlan1990learning, sloman1996empirical}, enabling the precise tracing of the specific path or rule that leads to a particular conclusion.
This characteristic facilitates the establishment of proofs and verification processes, thereby ensuring that derived statements are sound based on the given axioms~\cite{sloman1996empirical}. 
In contrast, LLMs, as neural-based models, often act as ``black boxes,'' introducing a level of unconstrained behavior that poses challenges in following strict logical reasoning~\cite{min2023recent}. To enhance systematic reasoning, various strategies were proposed to integrate LLMs with classical logical inference algorithms such as forward chaining~\cite{creswell2022selection} and backward chaining~\cite{kazemi2022lambada}. 
Yet, applying these techniques in open domains presents significant challenges as they frequently necessitate supplementary context or logical rules to provide constraints. Creating such logical rules can be demanding, especially with limited resources. Our proposed approach provides a systematic solution to address gaps in cases where explicit rules are absent. It achieves this by exploring the underlying graph structure of unstructured text, potentially improving the capability of Language Models (LLMs) to effectively traverse reasoning paths.

\subsection{Multi-step Reasoning over KGs}
KGs provide an effective way to explicitly organize information in the form of a structured graph. Multi-step reasoning naturally aligns with graph-based techniques, utilizing explicit pathways in the graph to represent the reasoning process~\cite{zhang2021neural,chen2020review}. 
For example, multi-step reasoning has been formulated in a reinforcement learning setup, where a policy-based agent sequentially extends its inference path until it reaches a target~\cite{das2018minerva, shen2018m, xiong2017deeppath, lin2018multi}. 
Moreover, to address the challenge of the more complex logical query answering in KGs, the query embedding method is proposed to conduct complex logical reasoning in the embedding space~\cite{GQE,ren2020query2box, ren2020beta}. This method involves transforming a First-Order Logic (FOL) query into a vector within the embedding space and subsequently searching for entities in the KG that share similar embeddings. Despite significant efforts to use KGs for direct reasoning, these graphs are often domain-specific and suffer from data sparsity. This means they might not have enough information for accurate multi-step reasoning across various topics. LLMs, on the other hand, can access a vast range of unstructured text, offering broader knowledge and topic coverage. To combine the strengths of both KGs and LLMs, there have been attempts to use KGs as external tools to incorporate additional facts into the reasoning process~\cite{pan2024unifying}. For instance, MindMap~\cite{wen2023mindmap} uses KGs to provide LLMs with up-to-date information and help them find reasoning paths. However, constructing and maintaining these KGs can be expensive and might even overwhelm LLMs with too much irrelevant information when addressing specific queries. In contrast, our approach takes a distinctive route. We hold the belief that natural language paragraphs inherently contain sufficient information for answering questions effectively. Instead of depending on external KGs, our proposition involves refining the organization of information within these paragraphs to enhance information retrieval and reasoning.


%% file: content/method.tex
\section{Framework: \model{}}\label{sec:Frameworks}
We propose \textit{\model{}}, \textit{a zero-shot prompting framework to guide LLMs in multi-step reasoning by explicitly converting unstructured text into a graph and instructs LLMs to navigate this graph using task-specific strategies to formulate responses.} 
The proposed framework is general, inherently task-agnostic, and capable of eliciting multi-step reasoning across broader natural language scenarios with a unified template. 

\subsection{Three-stage prompting} 
Our three-stage prompting method, inspired by human problem-solving with graphs, involves: {\color{olive}(1) Generating a graph from the given context;} {\color{teal}(2) Planning how to navigate the graph considering the tasks;} {\color{violet}(3) Executing the plan by traversing the graph to find the answer.} This approach mirrors how humans tackle graph-based problems. To facilitate recognition, each stage of the prompt is color-coded: {\color{olive}olive for the first stage}, {\color{teal}teal for the second}, and {\color{violet}violet for the third}.

\textbf{Example}
Let's illustrate the three-stage prompting using an example. Consider the following paragraph: \textit{``Christian got his son, \textbf{{\color{red}Seth}}, a car for his birthday. Christian and his brother Jonathan went to a basketball game. Jonathan's sister Ruth decided to tag along with them. Ruth invited her daughter Stephanie to lunch. Stephanie's brother \textbf{{\color{red}Jeremy}} couldn't leave work to join them.''} The question is to determine the family relationship between \textit{Seth} and \textit{Jeremy}.

\begin{figure}[h]
    \centering
    \includegraphics[width= \linewidth]{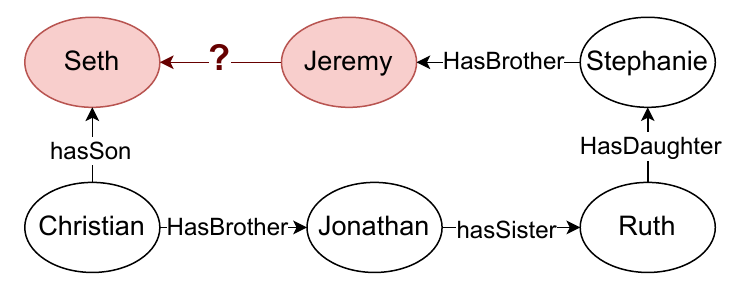}
    \caption{The graph representation of a story from CLUTRR dataset. our objective is to determine the family relationship between two nodes, \textit{Seth} and \textit{Jeremy}, which are highlighted in red.}
    \label{fig:graph representation}
\end{figure}

\textbf{{\color{olive}1st stage prompt: Concept Map Construction}}
In the first step, our goal is to convert an unstructured paragraph into a structured graph. Within this graph, each node corresponds to an entity, and the interconnecting edges depict the relationships linking these entities. Consider the given example, its graph representation is given in the Fig.~\ref{fig:graph representation}. 

\textbf{{\color{teal}2nd stage prompt: Task-specific Planning}} 
Fig.~\ref{fig:graph representation} demonstrates that,  while the text and question seem straightforward, the graph reveals a complex path between \textit{Seth} and \textit{Jeremy}. Correctly navigating this path requires specific planning strategies that direct the reasoning process.
It is important to note that these planning strategies are generally independent of the underlying data. The choice of strategy is task-specific. Consider the given example, to identify missing relationships between two entities (i.e., (\textit{Seth}, ? ,\textit{Jeremy})), our method involves deducing this link by tracing a path between them. We start with the subject entity and iteratively explore the most relevant information, progressing step-by-step until we reach the object entity. From there, we deduce the missing relation by analyzing the path between the two entities. To enhance the versatility of our framework, the following section will discuss different planning strategies applicable for a range of tasks.

\textbf{{\color{violet}3rd stage prompt: Execution with the Concept Map}}
Upon defining the task-specific planning strategy, we proceed to the execution phase, leveraging the concept map developed in 1st stage. This phase of instantiation enables us to address the specific problem in the given context. As illustrated in  Fig~\ref{fig:graph representation}, to derive the answer, we traverse the graph following the plan and carry out inference step by step:
Step 1:
Given $\text{Seth}\xleftarrow{\text{hasSon}}\text{Christian}\xrightarrow{\text{hasBrother}}\text{Jonathan}$, we have $\text{Seth}\xrightarrow{\text{hasUncle}}\text{Jonathan}$; Step 2:
Given $\text{Seth}\xrightarrow{\text{hasUncle}}\text{Jonathan} \xrightarrow{\text{hasSister}}\text{Ruth}$, we have $\text{Seth}\xrightarrow{\text{hasAunt}}\text{Ruth}$; Step 3:
Given $\text{Seth}\xrightarrow{\text{hasAunt}}\text{Ruth} \xrightarrow{\text{hasDaughter}}\text{Stephanie}$, we have $\text{Seth}\xrightarrow{\text{hasCousin}}\text{Stephanie}$; Step 4:
Given $\text{Seth}\xrightarrow{\text{hasCousin}}\text{Stephanie} \xrightarrow{\text{hasBrother}}\text{Jeremy}$, we have $\text{Seth}\xrightarrow{\text{hasCousin}}\text{Jeremy}$. 

\textbf{A Complete Prompt.}
By combining all three stages, we present the complete prompt for the given example: {\color{olive}First, create a knowledge graph by extracting facts from each sentence in the given input story.} {\color{teal}Once this is done, I will pose a question. This question can be transformed into a triple (s, ?, o), where your primary task is to determine the missing relation (‘?’) that links the subject entity (‘s’) to the object entity (‘o’). To begin, focus on the subject entity in this triple and choose the most relevant facts to expand from it. Step by step, progress towards the object entity, ensuring that each selected fact contributes to creating a link between the subject and object entities.} {\color{violet}Finally, utilize the established connection between the subject and object entities to answer the question.} 

%% file: content/tasks.tex
\section{Exploring Representative KG Reasoning Tasks}\label{sec:tasks}
Our framework is inherently task-agnostic, designed to accommodate a wide range of tasks with versatility. To cater to this diversity, we establish task-specific planning in 2nd stage prompt, tailored to each unique task. This section outlines various planning approaches for different tasks, demonstrating the framework's adaptability. We have provided all these prompts in Appendix A.2.

\subsection{Relation Prediction}
Relation prediction is a task focused on predicting the missing relations between two given entities, represented as $(h, ?, t)$. This task typically involves inferring the missing relations by tracing the path that links the target entities (i.e., $h$ and $t$) within the graph. We have discussed the planning strategies applicable to relation prediction task in Sec.~\ref{sec:Frameworks}.


\subsection{Entity Prediction}
Entity prediction is a fundamental task in KGs that aims to infer the missing entity in a given query, such as $(h, r, ?)$ or $(?, r, t)$. For example, the question ``Who currently holds the position of President in the USA?'' can be structured as a link prediction task within a KG, seeking to resolve the query $(?, \text{isPresidentOf}, \text{USA})$. This query could be straightforward, in this paper, we focus on more complex queries which require multi-step inference across various natural language scenarios.

\subsubsection{Entity Prediction over Dynamic KG}
Given that the information in Knowledge Graphs (KGs) can change dynamically, each time step introduces new information for inference. Therefore, predicting entities within dynamic KGs necessitates a step-by-step understanding of the status at each time interval to effectively manage entity prediction. For instance, consider the scenario: \textit{``Alice, Bob, and Claire are holding a white elephant gift exchange. At the start of the event, they are each holding a present of a different color: Alice has a yellow present, Bob has a brown present, and Claire has a blue present. As the event progresses, pairs of people swap gifts. \textbf{First}, Bob and Alice swap their gifts. \textbf{Then}, Claire and Alice swap their gifts. \textbf{Finally}, Bob and Alice swap their gifts. At the end of the event, what color gift does Bob have?''} This situation exemplifies entity prediction over dynamic KG, where the query can be structured as 
$(\text{Bob}, \text{hasGift}, ?)$. To accurately reflect the status at the event's conclusion, it is essential to capture changes at every time step, considering that each change depends on the previous time step. The primary planning strategy involves systematically tracking and recording the sequence of changes, with the KG at time step $t$ being modified based on the KG at the previous time step $t-1$.

\subsubsection{Complex Entity Prediction}
\begin{figure}[h]
    \centering
    \includegraphics[width=\linewidth]{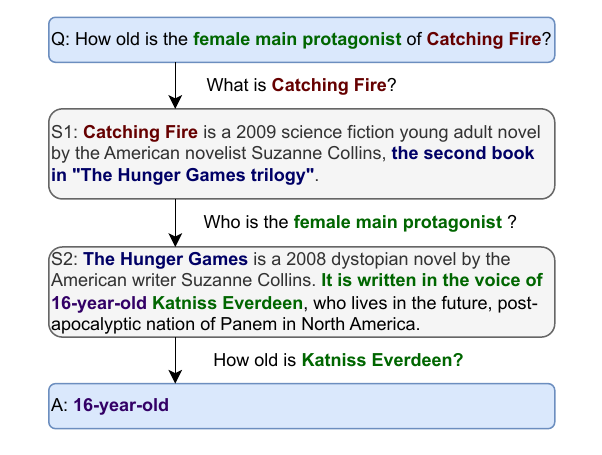}
    \caption{The bridging question in HotpotQA. It relies on multi-hop sequential reasoning to answer the question.}
    \label{fig:Complex Query Answering}
\end{figure}

While the previous method targets simpler one-hop queries in the form of $(h, r, ?)$, complex entity prediction aims to predict answers for queries with a more complex structure. For example, ``Riom Trial was headed by the French general who reached what distinction?'' is a complex query. This question's complexity goes beyond a straightforward relation, resembling a formal logic expression: $V_? := (\text{Riom Trial}, \text{wasHeadedBy}, V) \land (V, \text{Reached}, V_?)$. The bridging questions in HotpotQA provide typical examples of complex entity prediction, as illustrated in Fig.~\ref{fig:Complex Query Answering}. 
Answering these requires aggregating and linking data from disparate sections of a text, following a specific sequence to construct the final answer. The primary planning strategy involves decomposing the question into simpler sub-questions and tackling these sub-questions sequentially, referencing the knowledge graph for information.

\subsection{Graph Sorting}
Graph sorting task involves organizing entities within a graph according to a specified sequence. For instance, consider the scenario: \textit{``The following paragraphs each describe three objects arranged in a fixed order. The statements are logically consistent within each paragraph. On a branch, there are three birds: a blue jay, a quail, and a falcon. The falcon is to the right of the blue jay. The blue jay is to the right of the quail. Which bird is the second one counting from the left?''} To solve this, the main approach is to arrange the birds in the correct order based on the given information and then determine the answer.

\subsection{Graph Query}
Graph query task involves specifying a condition to retrieve specific data from a graph. For instance, consider the scenario: \textit{``Here is a table where the first line is a header and each subsequent line is a penguin:  name, age, height (cm), weight (kg) Louis, 7, 50, 11 Bernard, 5, 80, 13 Vincent, 9, 60, 11 Gwen, 8, 70, 15  For example: the age of Louis is 7, the weight of Gwen is 15 kg, the height of Bernard is 80 cm. How many penguins are more than 5 years old?''} This type of query can be expressed in SPARQL, a query language for databases, as follows:
\begin{verbatim}
SELECT (COUNT(?penguin) AS ?count)
WHERE {
    ?penguin ex:age ?age .   
    FILTER (?age > 5)      
}
\end{verbatim}
The primary planning strategy involves identifying the condition and selecting the entities that meet the condition from the graph.

\subsection{Logical Inference}
Logical inference and entailment are fundamental concepts in logic and reasoning, used in various fields to determine the logical relationships between statements. For instance, consider the scenario: \textit{``Sentence 1: as the mass of a celestial object decreases, the surface gravity of that celestial object weakens. Sentence 2: less is the opposite of more. Sentence 3: as the force of gravity decreases, the weight of the object will decrease. Sentence 4: an astronaut is a kind of object. Sentence 5: The Earth has more mass than the Moon. Sentence 6: surface gravity is a kind of force of gravity. Why do astronauts weigh more on Earth than they do on the Moon?''} To answer this, we construct a logical sequence: Astronauts experience greater weight on Earth than on the Moon due to Earth's stronger gravitational force. This is inferred from the fact that Earth, having more mass than the Moon, exerts a stronger surface gravity. The primary planning strategy involves beginning with the subject entities referenced in the question and establishing a logical chain based on the provided context.

%% file: content/experiment.tex
\section{Results}

For each task, we evaluate the performance of two LLM models, GPT-4 (\textit{gpt-4}) and GPT-3.5 (\textit{gpt-3.5-turbo})~\cite{OpenAI2023GPT4TR}. Since both methods are closed-source, we do not have specific information about their size, architecture, and pretraining particulars. For every task, we conduct a comparative analysis of our prompting framework against both with and without zero-shot chain-of-thought prompt (0-CoT), where 0-CoT encourages the model to engage in step-by-step reasoning by incorporating the phrase ``Let's think step by step'' in the prompts. We include the prompts for all six tasks in Appendix~\ref{app:Prompts} for reference.


\begin{figure*}[ht]
    \centering
    \begin{minipage}{0.49\textwidth}
        \centering
        \includegraphics[width=0.49\linewidth]{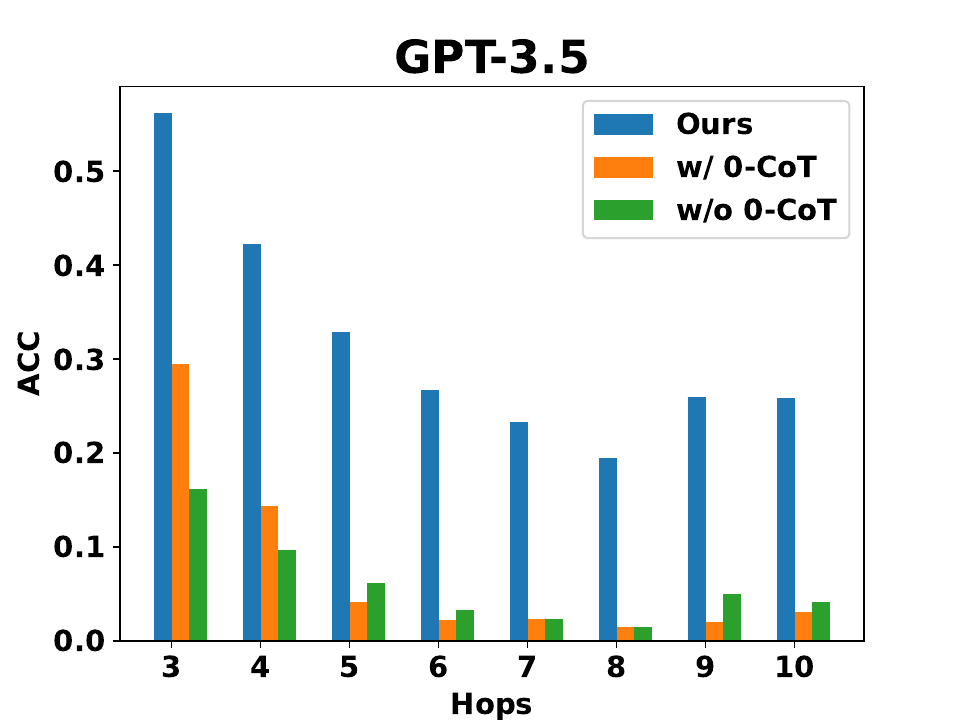}
        \hfill
        \includegraphics[width=0.49\linewidth]{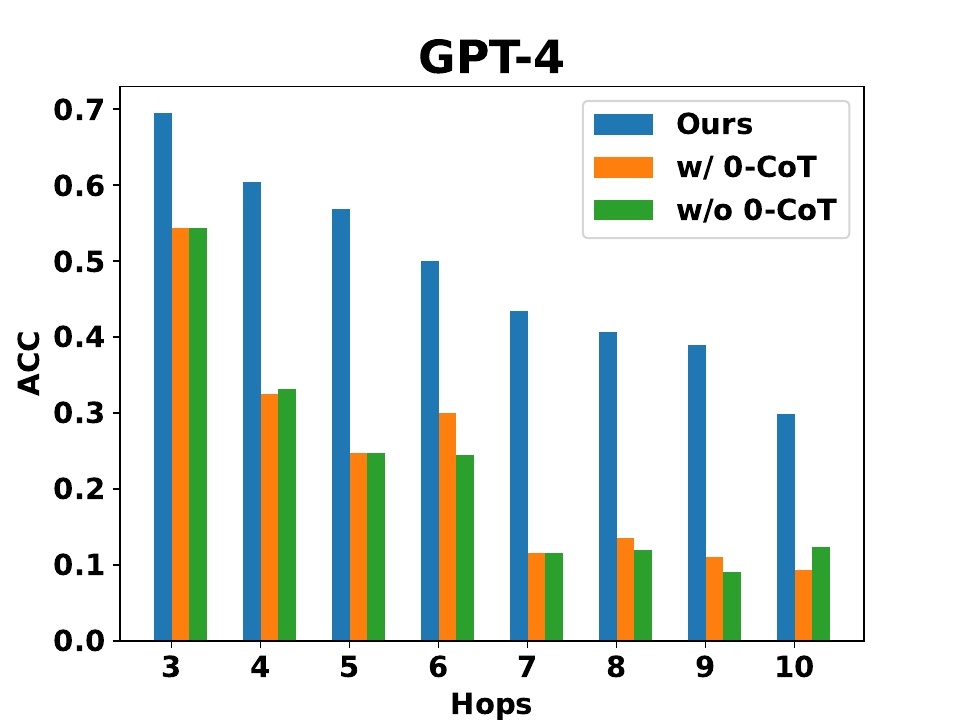}
        \caption*{Relation Prediction}
    \end{minipage}
    \hfill
    \begin{minipage}{0.49\textwidth}
        \centering
        \includegraphics[width=0.49\linewidth]{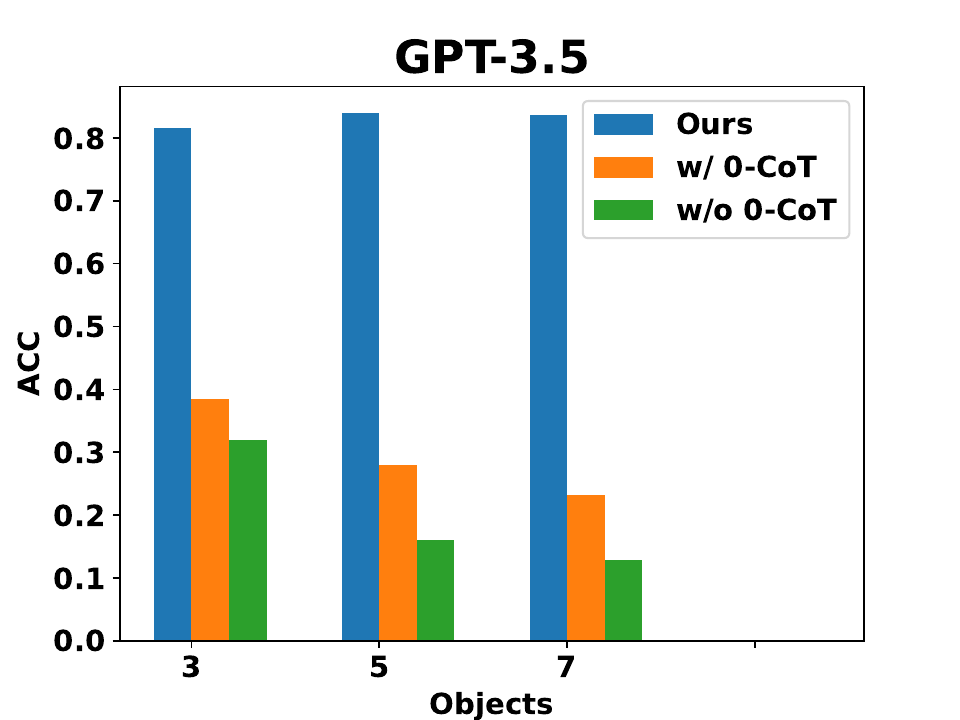}
        \hfill
        \includegraphics[width=0.49\linewidth]{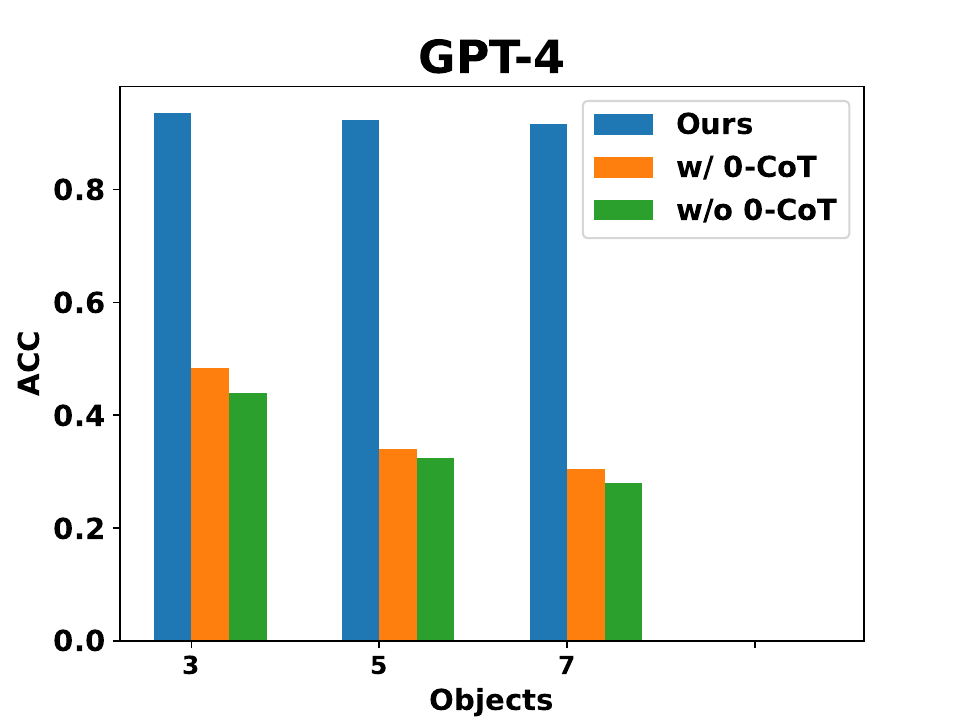}
        \caption*{Entity Prediction over Dynamic KG}
    \end{minipage}
    \hfill
    \begin{minipage}{0.49\textwidth}
        \centering
        \includegraphics[width=0.49\linewidth]{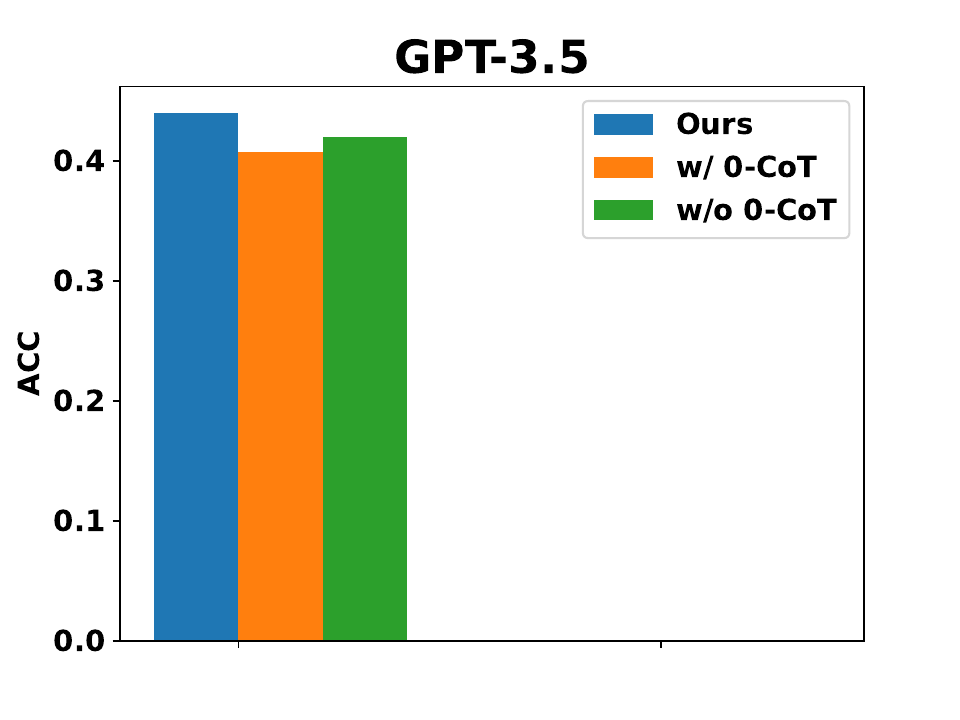}
        \hfill
        \includegraphics[width=0.49\linewidth]{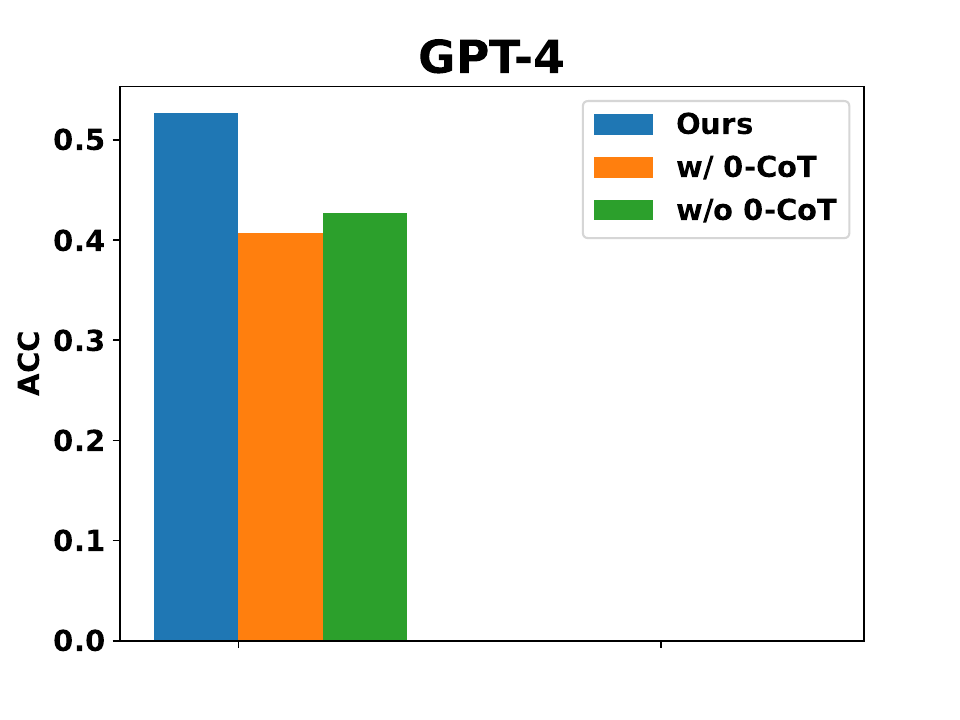}
        \caption*{Complex Entity Prediction}
    \end{minipage}
    \hfill
    \begin{minipage}{0.49\textwidth}
        \centering
        \includegraphics[width=0.49\linewidth]{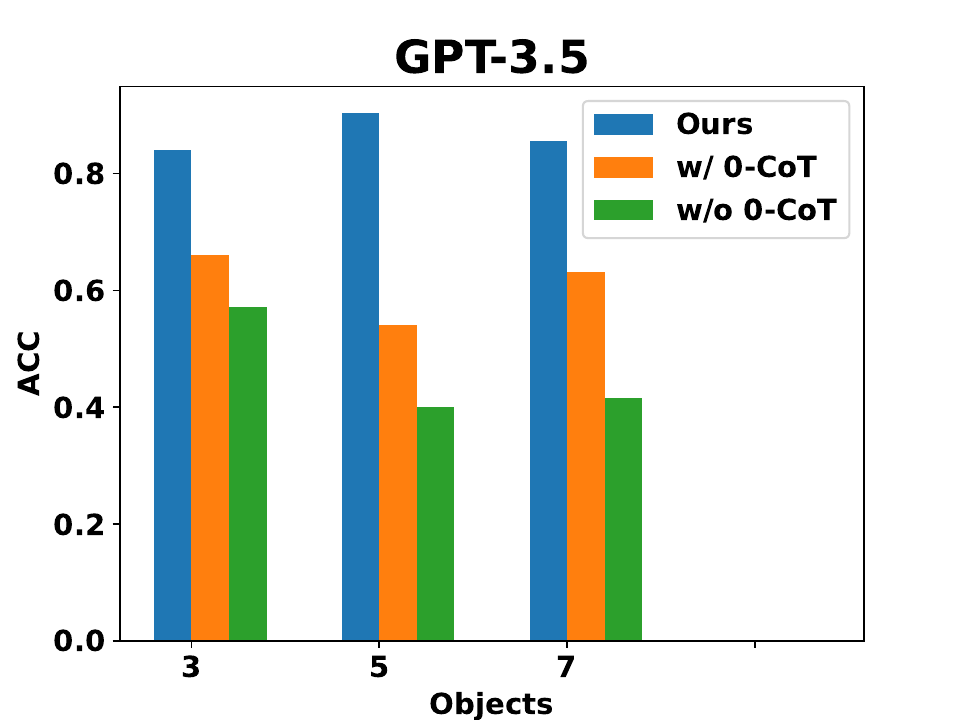}
        \hfill
        \includegraphics[width=0.49\linewidth]{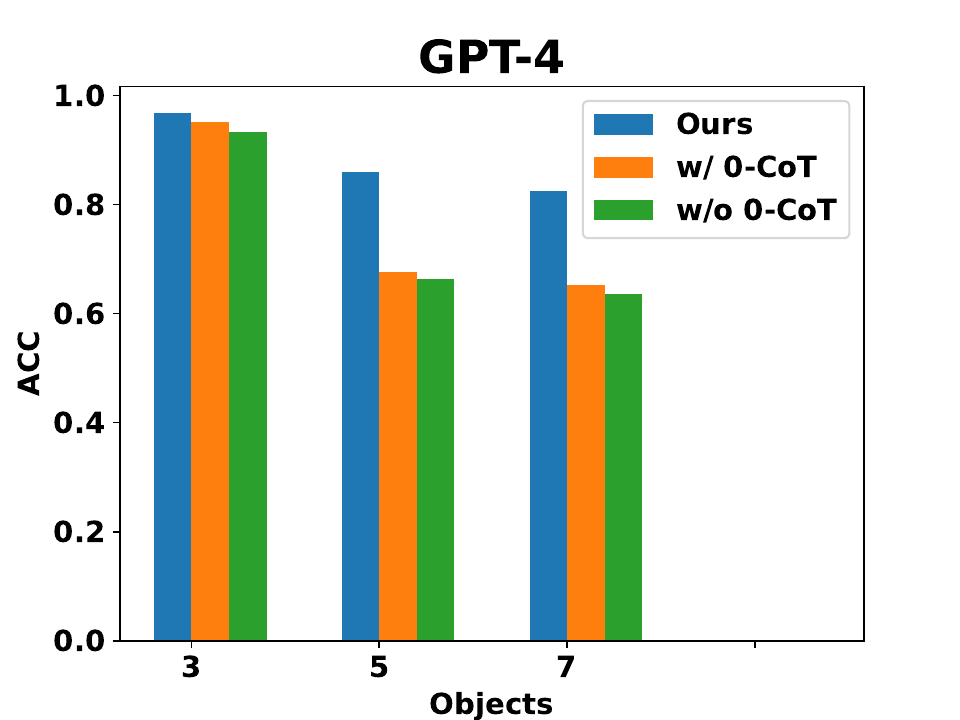}
        \caption*{Graph Sorting}
    \end{minipage}
    \begin{minipage}{0.49\textwidth}
        \centering
        \includegraphics[width=0.49\linewidth]{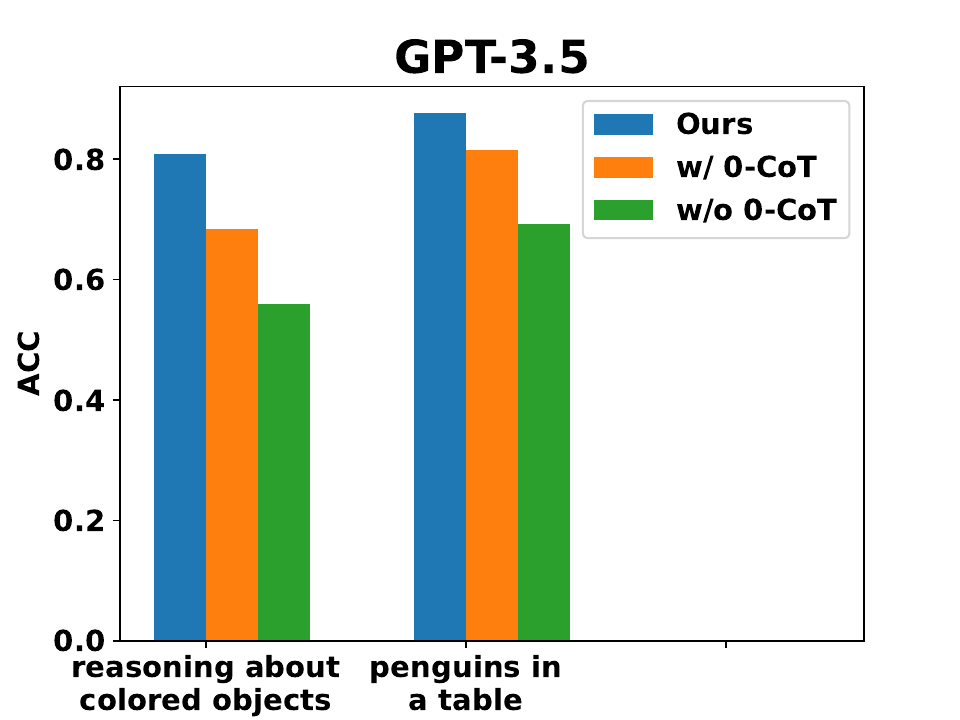}
        \hfill
        \includegraphics[width=0.49\linewidth]{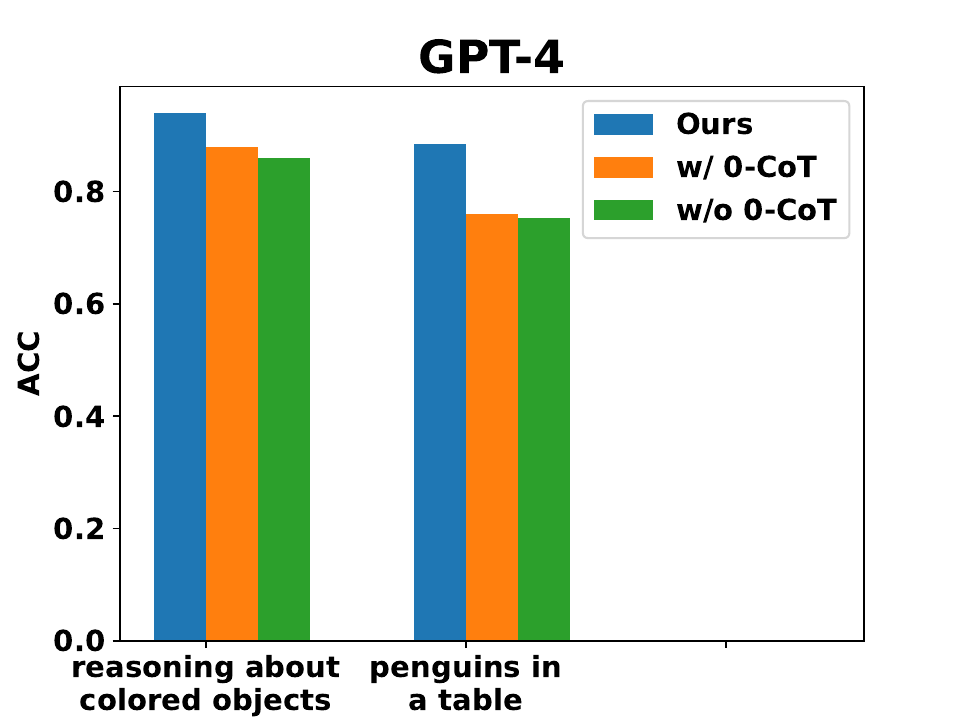}
        \caption*{Graph Query}
    \end{minipage}
    \hfill
    \begin{minipage}{0.49\textwidth}
        \centering
        \includegraphics[width=0.49\linewidth]{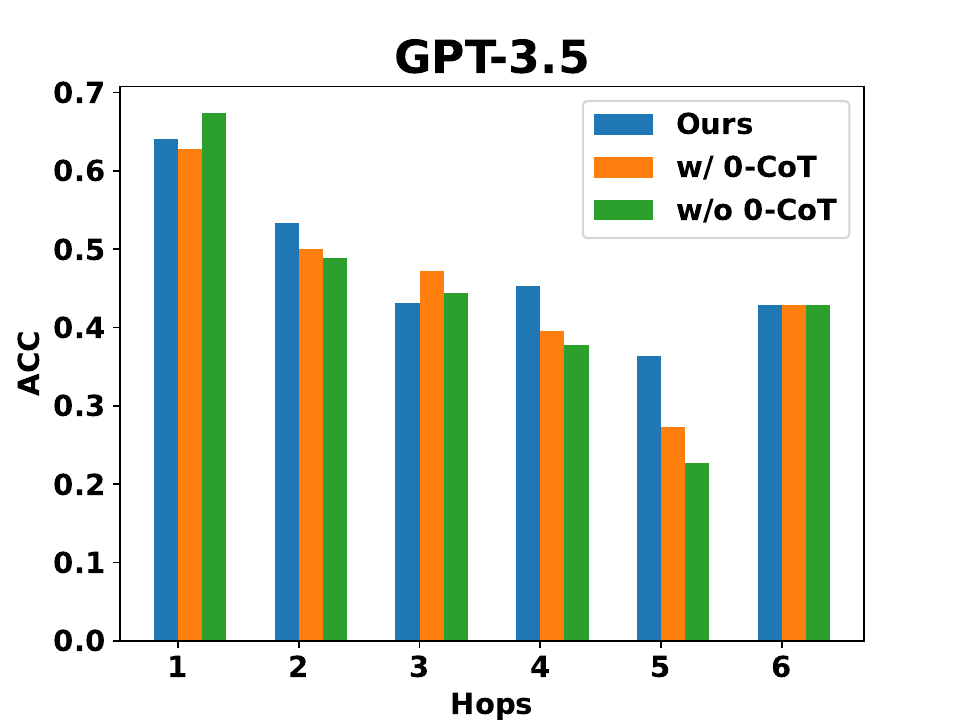}
        \hfill
        \includegraphics[width=0.49\linewidth]{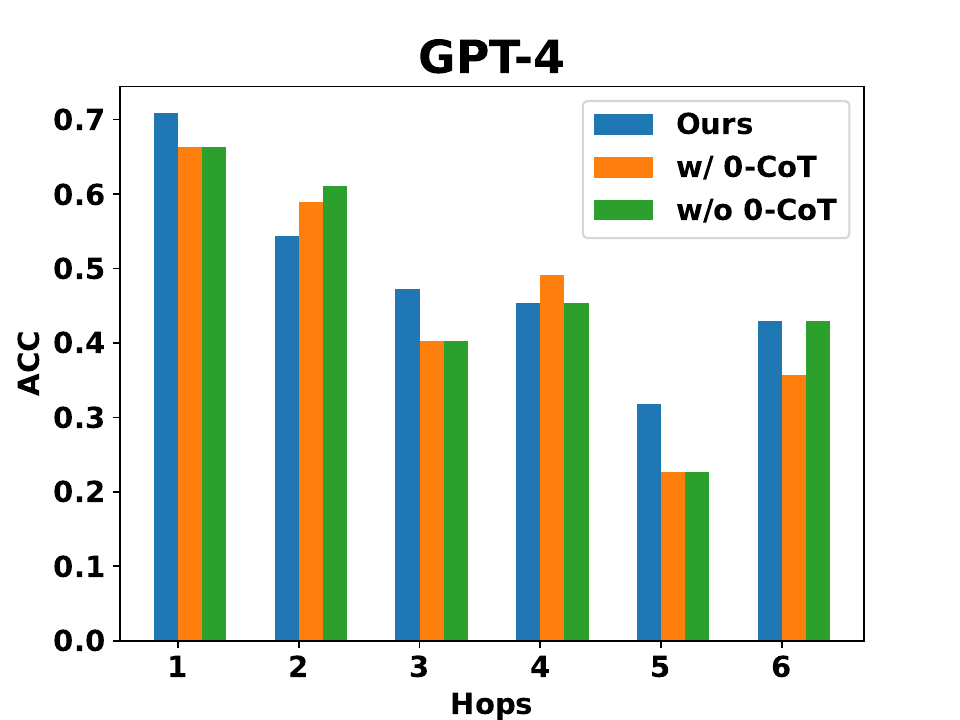}
        \caption*{Logical Inference}
    \end{minipage}
\caption{Main results. Different methods are illustrated through color-coded bars: {\color{blue}blue} bars indicate the results achieved using our {\color{blue}\textit{\model{}}}, while {\color{orange}orange} bars show the performance {\color{orange}with 0-shot chain-of-thought(0-CoT)}. Additionally, {\color{green}green} bars depict the performance {\color{green}without 0-CoT}. These results demonstrate that the \textit{\model{}} consistently and significantly outperforms the other methods, both with and without 0-CoT, across GPT-3.5 and GPT-4 models.}
\label{fig:results}
\end{figure*}

\subsection{Dataset}
We have incorporated four datasets:  \textit{CLUTRR}~\cite{sinha2019clutrr}, \textit{BIG-bench-hard (BBH)}~\cite{suzgun2022challenging}, \textit{HotpotQA}~\cite{yang2018hotpotqa} and \textit{Entailment Bank}~\cite{dalvi2021explaining} in experiments. These datasets cover all six tasks discussed in Sec.~\ref{sec:tasks}. Please refer to the Appendix~\ref{app:datasets} for detailed information.
\subsection{Analysis}\label{sec:Analysis}
\textbf{Relation Prediction}
The most representative dataset for relation prediction task is \textit{CLUTRR}~\cite{sinha2019clutrr}. It is a benchmark designed to infer the missing relationship between two individuals within a family network. To assess the complexity of the questions within the \textit{CLUTRR}, we have organized them based on the length of the relational paths connecting the target family members, typically spanning 3 to 10 hops. As shown in Fig.~\ref{fig:results}, this task poses a formidable challenge for LLM, even when the path length is relatively short. Even with the utilization of GPT-4 with 0-CoT, satisfactory performance remains elusive. This observation underscores the inherent limitations of LLM in handling datasets with significant relational complexity. Furthermore, as the length of the relational paths extends, the challenge intensifies. In contrast, with our proposed \textit{\model{}}, we can observe that it drastically increases the performance and suffers from less performance degradation when the path length increases.

\textbf{Entity Prediction over Dynamic KG}: 
We have included the \textit{tracking shuffled objects} datasets from BBH~\cite{suzgun2022challenging} to assess the entity prediction performance within dynamic KG. These datasets are designed to infer the relative positions of various shuffled objects at the conclusion of a narrative. The questions are organized according to the number of objects involved. As shown in Fig.~\ref{fig:results}, this task presents a significant challenge for LLM due to the requirement of maintaining an evolving graph representation at each time step when tracking shuffled objects. Given our proposed \textit{\model{}}, which explicitly constructs and tracks changes within the KG, we can observe a remarkable performance enhancement (e.g., improve by 146\% over GPT-4 w/ 0-CoT).

\textbf{Complex Entity Prediction}
The bridging questions in \textit{HotpotQA}~\cite{yang2018hotpotqa} provide typical examples for complex entity prediction task. As shown in Fig.~\ref{fig:results}, while our proposed \textit{\model{}} enhances performance, the improvement is not as significant as in other tasks. This is because the paragraphs in \textit{HotpotQA} are exceptionally long, making it challenging for LLM to construct a KG that encompasses every piece of information within the context. Consequently, our proposed \textit{\model{}} faces difficulty in further enhancing performance, especially with missing triples in the KG.

\textbf{Graph Sorting}
In the graph sorting task, we have included the \textit{logical deduction} datasets from BBH~\cite{suzgun2022challenging}. These datasets require to sort objects arranged in a line. The questions are organized according to the number of objects involved. As shown in Fig.~\ref{fig:results}, even though LLM already delivers impressive performance on this task, our proposed \textit{\model{}} brings about further improvements, particularly as the number of involved objects increases.

\textbf{Graph Query}
Within the graph query task, we have included \textit{reasoning about colored objects} and \textit{penguins in a table} datasets from BBH~\cite{suzgun2022challenging}. These datasets involve the selection and counting of objects that meet specific criteria from a given set of objects.  As shown in Fig.~\ref{fig:results}, LLMs already deliver impressive performance on this task, but our proposed \textit{\model{}} enhances performance even further.

\textbf{Logical Inference}: \textit{Entailment Bank}~\cite{dalvi2021explaining} is a widely used dataset for multi-step entailment tasks involving logical reasoning. To assess the complexity of the questions within the \textit{Entailment Bank}, we have categorized them based on the number of entailment steps required to arrive at an answer. As shown in Fig.~\ref{fig:results}, our proposed \textit{\model{}} doesn't consistently improve performance for this task. The challenge lies in the fact that logical reasoning often demands a precise order when constructing the logical graph, with rules typically dictating a direction from the premise to the conclusion. While following a forward chaining algorithm, one can readily employ the rules for logical inference sequentially, starting from known facts. However, in our scenario, we mix rules with facts and don't clearly distinguish between premises and conclusions within the rules. Consequently, even though we require LLMs to construct the logical graph, it remains challenging for LLMs to identify the correct logical order.

\subsection{Case Study}
In this subsection, we aim to highlight the advanced capabilities and improvements our model, \textit{\model{}}, offers over existing methodologies. We select the task of entity prediction over dynamic KG as a case for demonstration. Additional case studies covering other tasks can be found in the Appendix~\ref{app:case study}. 

\subsubsection{Case Study of Success}
We present a particular example where our \textit{\model{}} delivers an accurate prediction, underlining its precision and reliability, especially when compared to the erroneous prediction made by 0-CoT in the same context.

\underline{\textbf{Question:}} Alice, Bob, and Claire are friends and avid readers who occasionally trade books. At the start of the semester, they each buy one new book: Alice gets Hound of the Baskervilles, Bob gets The Pearl, and Claire gets The Odyssey.
As the semester proceeds, they start trading around the new books. First, Bob and Claire swap books. Then, Claire and Alice swap books. Finally, Bob and Alice swap books. At the end of the semester, Alice has Options:

(A) Hound of the Baskervilles

(B) The Pearl

(C) The Odyssey

\underline{\textbf{0-CoT Answer:}} (B) The Pearl \coloredcross

\underline{\textbf{\textit{\model{}} Answer:}} (C) The Odyssey

Explanation: Let's create the knowledge graph by extracting facts from each sentence in the story.

1. Alice, Bob, and Claire are friends and avid readers who occasionally trade books. 

   - (Alice, is\_friend\_with, Bob)
   
   - (Alice, is\_friend\_with, Claire)
   
   - (Bob, is\_friend\_with, Claire)
   
   - (Alice, is\_avid\_reader, True)
   
   - (Bob, is\_avid\_reader, True)
   
   - (Claire, is\_avid\_reader, True)
   
   - (Alice, trades\_books, True)
   
   - (Bob, trades\_books, True)
   
   - (Claire, trades\_books, True)

2. At the start of the semester, they each buy one new book: Alice gets Hound of the Baskervilles, Bob gets The Pearl, and Claire gets The Odyssey.

   - (Alice, has\_book, Hound of the Baskervilles)
   
   - (Bob, has\_book, The Pearl)
   
   - (Claire, has\_book, The Odyssey)

3. As the semester proceeds, they start trading around the new books. First, Bob and Claire swap books. 

   - (Bob, has\_book, The Odyssey)
   
   - (Claire, has\_book, The Pearl)

4. Then, Claire and Alice swap books. 

   - (Claire, has\_book, Hound of the Baskervilles)
   
   - (Alice, has\_book, The Pearl)

5. Finally, Bob and Alice swap books. 

   - (Bob, has\_book, The Pearl)
   
   - (Alice, has\_book, The Odyssey)

So, at the end of the semester, Alice has The Odyssey. Therefore, option (C) is correct. \coloredcheckmark

\underline{\textbf{Analysis:}}  By explicitly converting unstructured text into a graph and directing LLMs to monitor the evolution of this dynamic graph, we can observe that LLMs accurately discern the altered object even after numerous iterations.

\subsubsection{Case Study of Failure}
To comprehensively understand the limitations of our proposed methods, we provide a specific example demonstrating a scenario where our \textit{\model{}} yields an incorrect prediction, in contrast to the accurate prediction made by 0-CoT under the same circumstances.

\underline{\textbf{Question:}} Alice, Bob, and Claire are dancers at a square dance. At the start of a song, they each have a partner: Alice is dancing with Rodrigo, Bob is dancing with Jamie, and Claire is dancing with Lola. Throughout the song, the dancers often trade partners. First, Claire and Alice switch partners. Then, Bob and Claire switch partners. Finally, Claire and Alice switch partners. At the end of the dance, Alice is dancing with Options:

(A) Rodrigo

(B) Jamie

(C) Lola

\underline{\textbf{0-CoT Answer:}} (B) Jamie \coloredcheckmark

\underline{\textbf{\textit{\model{}} Answer:}} (C) Lola

Explanation: 

Let's create the knowledge graph:

1. Alice, Bob, and Claire are dancers at a square dance. 

   - (Alice, is a dancer at, square dance)
   
   - (Bob, is a dancer at, square dance)
   
   - (Claire, is a dancer at, square dance)

2. At the start of a song, they each have a partner: Alice is dancing with Rodrigo, Bob is dancing with Jamie, and Claire is dancing with Lola.

   - (Alice, is dancing with, Rodrigo)
   
   - (Bob, is dancing with, Jamie)
   
   - (Claire, is dancing with, Lola)

3. First, Claire and Alice switch partners.

   - (Alice, is dancing with, Lola)
   
   - (Claire, is dancing with, Rodrigo)

4. Then, Bob and Claire switch partners.

   - (Bob, is dancing with, Rodrigo)
   
   - (Claire, is dancing with, Jamie)

5. Finally, Claire and Alice switch partners.

   \hl{- (Alice, is dancing with, Jamie)}\coloredcheckmark
   
   - (Claire, is dancing with, Lola)

\hl{So, at the end of the dance, Alice is dancing with Lola.} Therefore, option (C) is correct. \coloredcross

\underline{\textbf{Analysis:}} As we highlight, while the inference process of our proposed model is accurate, it makes mistakes during the conclusion-drawing phase. This pattern has been widely observed in our experiments.

\subsubsection{Conclusion} 
A notable finding is that while the LLMs successfully adheres to the prompts to construct accurate KGs and navigates these KGs correctly according to task-specific strategic guidance, it often makes mistakes during the conclusion-drawing phase, even with correct inference results immediately preceding this stage. This issue could potentially be addressed by employing an additional LLM to verify the consistency of the generated content. We plan to explore this approach to further improve our framework in the future.

%% file: content/discussion.tex
\section{Discussion}

\textbf{Do LLMs spontaneously represent natural language text as a KG for multi-step reasoning?}
LLMs like GPT-3 are mainly trained for predicting the next token based on context rather than structuring unstructured text into KGs for multi-step reasoning. Although LLMs are not naturally structured as KGs, they can be prompted for structured thinking. The ``Zero-Shot-CoT''~\cite{kojima2022large} approach, which prepends the input question with the phrase ``Let’s think step by step'' before querying the model, has shown promise in encouraging structured thinking and improving reasoning performance in LLMs. LLMs also excel in planning\cite{wang2023survey}, especially when breaking down complex questions into simpler sub-questions for sequential answers~\cite{zhou2022least, lee2023recursion, drozdov2022compositional}. 
However, while LLMs excel in generating sequential steps based on prompts, their effectiveness is limited when faced with real-world scenarios with a significant relational complexity as discussed in Sec.~\ref{sec:Analysis}. 

\textbf{Is a KG expressive enough to represent natural language text?}
KGs excel in structuring factual information and relationships, making them useful for organizing knowledge. However, the expressiveness of KGs can be limited when it comes to handling the richness of natural language. One major drawback is their inability to effectively convey emotions and sentiments. KGs are primarily designed for storing concrete information, making them less suitable for encoding human emotions. In contrast, natural language text allows for a broad range of emotional expressions, from joy to sorrow, humor to sarcasm. For example, the sentence ``If I were a bird, I would fly to far-off lands'' carries emotional weight and context-dependent meanings that KGs may struggle to capture. Depending on the context, it could express a longing for adventure, a desire for freedom, or metaphorically represent personal aspirations. These nuances are deeply embedded in natural language and not easily translatable into the rigid structure of a KG. While this paper demonstrates the value of KGs in promoting structured thinking in LLMs, it's important to recognize that KGs may have limitations in fully capturing the expressive power of natural language text. Further research is needed to bridge this gap.

\textbf{Do we have more effective methods for representing the relationships among various pieces of information?}
The question of more effective methods for representing relationships among information, especially in light of KGs' limitations in handling natural language, is vital. Expanding KGs to incorporate unary attributes alongside binary predicates to describe events' properties is beneficial. Moreover, it is crucial, as discussed in Sec.~\ref{sec:Analysis}, to enhance mechanisms for detecting textual entailment, contradiction, and inference. These enhancements can enable LLMs to more effectively participate in nuanced reasoning, including considerations of causality and temporal relationships. Given natural language's flexibility, seamless NLP pipelines that combine various models, such as named entity recognition, dependency parsing, and entity resolution, are essential. These unified pipelines may capture and clarify complex relationships within textual data, enhancing structured reasoning capabilities.

%% file: content/conclusion.tex
\section{Conclusion}
LLMs often excel in simple reasoning tasks but struggle with multi-step reasoning. Graphs offer an effective way to model relational data and capture long-term dependencies among entities. This paper bridges this gap by introducing an innovative task-agnostic prompting framework, \textit{\model{}}. This framework enhances the multi-step reasoning capabilities of LLMs within a zero-shot setting by systematically converting unstructured text into a graphical format and guiding LLMs in traversing this graph using task-specific strategies to construct responses. Our experiments show that our proposed framework significantly enhances the reasoning capabilities of LLMs, empowering them to excel in a broader spectrum of natural language scenarios.

%% file: content/appendix.tex
\clearpage
\appendix
\section{Appendix}
\subsection{Datasets}\label{app:datasets}
\textbf{CLUTRR}
The most representative dataset employed for \textit{relation prediction} is CLUTRR~\cite{sinha2019clutrr} (Compositional Language Understanding with Text-based Relational Reasoning). It is a benchmark designed to infer the relationship between two family members, which isn't directly mentioned in the story. Successful performance on this task requires both extracting relationships between entities, as well as inferring the logical rules governing these relationships. To assess the complexity of the questions within the CLUTRR dataset, we have categorized them based on the length of the paths between the target family members. For a comprehensive overview of the statistics of the CLUTRR dataset, please refer to Table~\ref{tab:CLUTRR}.

\begin{table*}
\centering
\caption{Statistics of the CLUTRR Datasets.} 
\label{tab:CLUTRR}
\begin{threeparttable} 
\begin{tabular}{c|c|c|c|c|c|c|c|c}
\toprule[1.5pt] 
& 3 hop & 4 hop & 5 hop & 6 hop & 7 hop & 8 hop & 9 hop & 10 hop \\
\hline
\# queries & 105 &154 &146 &90 &129 &133 &100 &97
\\
\bottomrule[1.5pt] 
\end{tabular}
\end{threeparttable}
\end{table*}

\textbf{BIG-bench}
The Beyond the Imitation Game Benchmark (BIG-bench)~\cite{srivastava2022beyond} is a comprehensive dataset designed for evaluating the capabilities of LLMs. BIG-bench is notable for its emphasis on tasks that pose significant challenges to current AI models. This benchmark includes a wide variety of tasks that test various aspects of language understanding and generation. These tasks can range from simple arithmetic to complex reasoning, understanding of cultural references, and more. 
Among these, 23 particularly demanding tasks have been aggregated to form the BIG-Bench Hard (BBH)~\cite{suzgun2022challenging} dataset. These tasks were selected because prior language model evaluations did not surpass average human performance on them. Several tasks within BBH provide typical examples that align with our research. For instance, the tracking shuffled objects tasks aligns with \textit{entity prediction over dynamic KG}. The logical deduction dataset is apt for \textit{graph sorting} task. Datasets like reasoning about colored objects and penguins in a table are suitable for \textit{graph query} task.

\textbf{HotpotQA}
HotpotQA~\cite{yang2018hotpotqa} is a widely used multi-hop question-answering dataset. Unlike traditional QA datasets where the answer can be found in a single passage, HotpotQA requires the system to gather and integrate information from several passages to answer a question correctly. A notable feature of HotpotQA is its inclusion of \textit{bridging questions}, which are structured to start with an initial fact located in one passage. The challenge is to leverage this piece of information to uncover and comprehend a related fact in a different passage, effectively forming a ``bridge'' that connects the initial fact to the final answer. By integrating the insights gained from the secondary passage with the initial information, we can effectively address the bridge question. These bridging questions thus serve as typical examples for \textit{complex entity prediction} task. For our analysis, we randomly selected a subset of 150 hard bridging questions from the development sets of HotpotQA.

\textbf{Entailment Bank}
Entailment Bank~\cite{dalvi2021explaining} is a key dataset extensively employed for \textit{logical reasoning} studies. This dataset includes structured entailment trees that represent complex entailment reasoning in a hierarchical format. These trees are designed to provide step-by-step logical reasoning process from premises to conclusion. To assess the complexity of the questions within the Entailment Bank, we have categorized them based on the number of entailment steps required to arrive at an answer. For a comprehensive overview of the statistics of the Entailment Bank datasets, please refer to Table~\ref{tab:Entailment Bank}. 
\begin{table*}
\centering
\caption{Statistics of the Entailment Bank Datasets.} 
\label{tab:Entailment Bank}
\begin{threeparttable} 
\begin{tabular}{c|c|c|c|c|c|c}
\toprule[1.5pt] 
& 1 hop & 2 hop & 3 hop & 4 hop & 5 hop & 6 hop \\
\hline
\# queries & 86 &90 &72 &53 &22 &14 \\
\bottomrule[1.5pt] 
\end{tabular}
\end{threeparttable}
\end{table*}

\subsection{Case Study over Each Tasks}\label{app:case study}
In this section, we showcase examples of both correct and incorrect responses, generated by \textit{GPT-4}, utilizing the prompts we designed for each task.

\subsubsection{Relation Prediction}
{\color{blue}First, we showcase an example accurately predicted by our proposed \model{}, in contrast to its incorrect prediction by 0-CoT}.

\underline{\textbf{Question:}} [Seth] and his wife [Laura] went to the beach to watch the fireworks on the 4th of July. [Rosa] took her father [Seth] to his doctor's appointment and afterward her brother [Jason] came over and visited for a while. [Jonathan] wanted to have a family cookout so he invited his brother [Christian]. [Christian] wanted to invite his other brother [Jason]. What is the family relationship between: [Laura] with [Jonathan]?

\underline{\textbf{0-CoT Answer:}} The story does not provide information on the family relationship between Laura and Jonathan. \coloredcross

\underline{\textbf{\model{} Answer:}} The family relationship between Laura and Jonathan is not explicitly mentioned in the story. However, we can infer some relationships based on the information given.

Knowledge Graph:

1. (Seth, wife, Laura)

2. (Seth, daughter, Rosa)

3. (Rosa, brother, Jason)

4. (Jonathan, brother, Christian)

5. (Christian, brother, Jason)

From the knowledge graph, we can see that Seth is married to Laura and Seth is the father of Rosa and Jason. Jonathan is the brother of Christian, who is also a brother of Jason. Therefore, Jonathan and Jason are brothers, making Seth Jonathan's father. Since Laura is Seth's wife, she would be Jonathan's mother. 

So, the family relationship between Laura and Jonathan is that Laura is Jonathan's mother. \coloredcheckmark

{\color{red}Next, we present an example where our proposed \model{} made an incorrect prediction, in contrast to the correct prediction made by 0-CoT.}

\underline{\textbf{Question:}} [Christian] was tired of his sister, [Ruth], complaining about his cooking so [Christian] complained to his father, [Seth]. [Ruth] and her daughter [Stephanie] went to see a movie yesterday and then got ice cream afterwards. [Christian] invited his father [Seth] and his brother [Stephen] to go on a weekend fishing trip. What is the family relationship between: [Stephanie] with [Stephen]?

\underline{\textbf{0-CoT Answer:}} [Stephanie] is [Stephen]'s niece. \coloredcheckmark

\underline{\textbf{\model{} Answer:}} To answer the question, we need to find the relation between Stephanie and Stephen. Let's start by creating a knowledge graph from the given story.

1. (Christian, sibling, Ruth)

2. (Christian, child, Seth)

3. (Ruth, child, Seth)

4. (Ruth, parent, Stephanie)

5. (Christian, sibling, Stephen)

6. (Stephen, child, Seth)

Now, let's start with Stephanie and find the most relevant facts to expand from it.

From fact 4, we know that Ruth is Stephanie's mother. 

From fact 3, we know that Ruth is also Seth's child. 

From fact 6, we know that Stephen is also Seth's child. 

\hl{So, Stephanie and Stephen are both children of Seth, which makes them siblings.} Therefore, the missing relation is 'sibling'. \coloredcross

\fbox{\parbox{0.9\linewidth}{
\textbf{Analysis:} As we highlight, the error in the output of our model comes from an inaccurate conclusion. Specifically, Stephanie should be categorized as Seth's grandchild, not as his child. Consequently, the correct relation should be ``niece.''}}

\subsubsection{Entity Prediction over Dynamic KG}
{\color{blue}First, we showcase an example accurately predicted by our proposed \model{}, in contrast to its incorrect prediction by 0-CoT.}

\underline{\textbf{Question:}} Alice, Bob, and Claire are friends and avid readers who occasionally trade books. At the start of the semester, they each buy one new book: Alice gets Hound of the Baskervilles, Bob gets The Pearl, and Claire gets The Odyssey.
As the semester proceeds, they start trading around the new books. First, Bob and Claire swap books. Then, Claire and Alice swap books. Finally, Bob and Alice swap books. At the end of the semester, Alice has Options:

(A) Hound of the Baskervilles

(B) The Pearl

(C) The Odyssey

\underline{\textbf{0-CoT Answer:}} (B) The Pearl \coloredcross

\underline{\textbf{\model{} Answer:}} (C) The Odyssey

Explanation: Let's create the knowledge graph by extracting facts from each sentence in the story.

1. Alice, Bob, and Claire are friends and avid readers who occasionally trade books. 

   - (Alice, is\_friend\_with, Bob)
   
   - (Alice, is\_friend\_with, Claire)
   
   - (Bob, is\_friend\_with, Claire)
   
   - (Alice, is\_avid\_reader, True)
   
   - (Bob, is\_avid\_reader, True)
   
   - (Claire, is\_avid\_reader, True)
   
   - (Alice, trades\_books, True)
   
   - (Bob, trades\_books, True)
   
   - (Claire, trades\_books, True)

2. At the start of the semester, they each buy one new book: Alice gets Hound of the Baskervilles, Bob gets The Pearl, and Claire gets The Odyssey.

   - (Alice, has\_book, Hound of the Baskervilles)
   
   - (Bob, has\_book, The Pearl)
   
   - (Claire, has\_book, The Odyssey)

3. As the semester proceeds, they start trading around the new books. First, Bob and Claire swap books. 

   - (Bob, has\_book, The Odyssey)
   
   - (Claire, has\_book, The Pearl)

4. Then, Claire and Alice swap books. 

   - (Claire, has\_book, Hound of the Baskervilles)
   
   - (Alice, has\_book, The Pearl)

5. Finally, Bob and Alice swap books. 

   - (Bob, has\_book, The Pearl)
   
   - (Alice, has\_book, The Odyssey)

So, at the end of the semester, Alice has The Odyssey. Therefore, option (C) is correct. \coloredcheckmark

{\color{red}Next, we present an example where our proposed \model{} made an incorrect prediction, in contrast to the correct prediction made by 0-CoT.} 

\underline{\textbf{Question:}} Alice, Bob, and Claire are dancers at a square dance. At the start of a song, they each have a partner: Alice is dancing with Rodrigo, Bob is dancing with Jamie, and Claire is dancing with Lola. Throughout the song, the dancers often trade partners. First, Claire and Alice switch partners. Then, Bob and Claire switch partners. Finally, Claire and Alice switch partners. At the end of the dance, Alice is dancing with Options:

(A) Rodrigo

(B) Jamie

(C) Lola

\underline{\textbf{0-CoT Answer:}} (B) Jamie \coloredcheckmark

\underline{\textbf{\model{} Answer:}} (C) Lola

Explanation: 

Let's create the knowledge graph:

1. Alice, Bob, and Claire are dancers at a square dance. 

   - (Alice, is a dancer at, square dance)
   
   - (Bob, is a dancer at, square dance)
   
   - (Claire, is a dancer at, square dance)

2. At the start of a song, they each have a partner: Alice is dancing with Rodrigo, Bob is dancing with Jamie, and Claire is dancing with Lola.

   - (Alice, is dancing with, Rodrigo)
   
   - (Bob, is dancing with, Jamie)
   
   - (Claire, is dancing with, Lola)

3. First, Claire and Alice switch partners.

   - (Alice, is dancing with, Lola)
   
   - (Claire, is dancing with, Rodrigo)

4. Then, Bob and Claire switch partners.

   - (Bob, is dancing with, Rodrigo)
   
   - (Claire, is dancing with, Jamie)

5. Finally, Claire and Alice switch partners.

   \hl{- (Alice, is dancing with, Jamie)}\coloredcheckmark
   
   - (Claire, is dancing with, Lola)

\hl{So, at the end of the dance, Alice is dancing with Lola.} Therefore, option (C) is correct. \coloredcross

\fbox{\parbox{0.9\linewidth}{
\textbf{Analysis:} As we highlight, while the inference process of our proposed model is accurate, it makes mistakes during the conclusion-drawing phase. This pattern has been widely observed in our experiments.}}

\subsubsection{Complex Entity Prediction}
{\color{blue}First, we showcase an example accurately predicted by our proposed \model{}, in contrast to its incorrect prediction by 0-CoT.}

\underline{\textbf{Question:}} Paragraph 0 The Lighthouse (Poole): The Lighthouse is an arts centre in Poole, Dorset, England.According to the Arts council of England it is the largest arts centre in the United Kingdom outside London.

Paragraph 1 Canford Cliffs: Canford Cliffs is an affluent suburb of Poole in Dorset, England.The neighbourhood lies on the English Channel coast midway between Poole and Bournemouth.To the southwest is Sandbanks which has some of the highest property values in the world, and together Canford Cliffs and Sandbanks form a parish, which has the fourth highest property prices in the world and second highest in the United Kingdom after London.

Paragraph 2 Viscount Trenchard: Viscount Trenchard, of Wolfeton in the County of Dorset, is a title in the Peerage of the United Kingdom.It was created in 1936 for Marshal of the Royal Air Force, Hugh Trenchard, 1st Baron Trenchard.He had already been created a Baronet, of Wolfeton in the County of Dorset, in the Baronetage of the United Kingdom in 1919 and Baron Trenchard, of Wolfeton in the County of Dorset, in 1930, also in the Peerage of the United Kingdom.His second son, the second Viscount, held junior ministerial positions from 1979 to 1983 in the Conservative administration of Margaret Thatcher.s of 2016 the titles are held by the latter's son, the third Viscount, who succeeded in 1987.In 2004 he replaced the recently deceased Lord Vivian as one of the ninety elected(by hereditary peers)hereditary peers that are allowed to remain in the House of Lords after the passing of the House of Lords Act 1999.Lord Trenchard sits on the Conservative benches.

Paragraph 3 Fire Radio: Fire Radio is a United Kingdom radio station broadcasting to Bournemouth, Poole, and Christchurch, Dorset, based in Southampton, Hampshire.

Paragraph 4 Dorset County Council election, 2013: An election to Dorset County Council took place on 2 May 2013 as part of the United Kingdom local elections.45 councillors were elected from 42 electoral divisions, which returned either one or two county councillors each by first-past-the-post voting for a four-year term of office.The electoral divisions were the same as those used at the previous election in 2009.No elections were held in Bournemouth or Poole, which are unitary authorities outside the area covered by the County Council.The election saw the Conservative Party maintain overall control of the council.

Paragraph 5 Viscount Wimborne: Viscount Wimborne, of Canford Magna in the County of Dorset, is a title in the Peerage of the United Kingdom.It was created in 1918 for Ivor Guest, 2nd Baron Wimborne.The Guest family descends from the engineer and businessman John Josiah Guest.On 14 August 1838 he was created a baronet, of Dowlais in the County of Glamorgan, in the Baronetage of the United Kingdom.He was succeeded by his eldest son, the second Baronet.In 1880 he was created Baron Wimborne, of Canford Magna in the County of Dorset, in the Peerage of the United Kingdom.On his death the titles passed to his eldest son, the second Baron.In 1910, four years before he succeeded his father, he had been raised to the Peerage of the United Kingdom in his own right as Baron Ashby St Ledgers, of Ashby St Ledgers in the County of Northampton.On his retirement as Lord-Lieutenant of Ireland in 1918 he was further honoured when he was made Viscount Wimborne, of Canford Magna in the County of Dorset, in the Peerage of the United Kingdom.His son, the second Viscount, represented Breconshire in the House of Commons.s of 2014 the titles are held by the latter's grandson, the fourth Viscount, who succeeded his father in 1993.

Paragraph 6 Lush (company): Lush Ltd. is a cosmetics retailer headquartered in Poole, Dorset, United Kingdom.The company was founded by Mark Constantine, a trichologist and Liz Weir, a beauty therapist.They met in a hair and beauty salon in Poole, England.A few years later, they decided to branch out and start their own business selling natural hair and beauty products.

Paragraph 7 Baron de Mauley: Baron de Mauley, of Canford in the County of Dorset, is a title in the Peerage of the United Kingdom.It was created in 1838 for the Whig politician the Hon. William Ponsonby, who had earlier represented Poole, Knaresborough and Dorset in the House of Commons.He was the third son of the 3rd Earl of Bessborough, an Anglo-Irish peer, and the husband of Lady Barbara Ashley-Cooper, one of the co-heirs to the ancient barony by writ of Mauley (or Maulay), which superseded the feudal barony the "caput" of which was at Mulgrave Castle, Yorkshire, which barony by writ had become extinct in 1415.His son, later the second Baron, sat as Member of Parliament for Poole and Dungarvon.s of 2009 the title is held by the latter's great-great-grandson, the seventh Baron, who succeeded his uncle in 2002.He is one of the ninety elected hereditary peers that remain in the House of Lords after the passing of the House of Lords Act 1999, and sits as a Conservative.He was elected in 2005 and thereby became the first hereditary peer having succeeded to his title after the House of Lords Act of 1999, to have obtained an elective hereditary peers seat in the House of Lords.As a descendant of the third Earl of Bessborough, Lord de Mauley is also in remainder to the earldom of Bessborough and its subsidiary titles.

Paragraph 8 List of cider producers in Dorset: This is a list of cider brands from Dorset in the United Kingdom.Although neighbouring county of Somerset is better known for its apple orchards, Dorset has an equally long orchard and cider tradition.Many Dorset cider producers are using traditional Dorset varieties of apple The Dorset Cider museum is based in Owermoigne, near Dorchester

Paragraph 9 Poole Methodist Church: Poole Methodist Church (also known as Poole High Street Methodist Church or The Spire) is a nineteenth-century Methodist church on Poole High Street in Dorset, England.An extension to the church was nominated for the 2016 Carbuncle Cup for "the ugliest building in the United Kingdom completed in the last 12 months".

What retailer in ABQ Uptown is headquarted in Poole, Dorset, United Kingdom?

\underline{\textbf{0-CoT Answer:}} The story does not provide information on any retailer in ABQ Uptown that is headquartered in Poole, Dorset, United Kingdom. \coloredcross

\underline{\textbf{\model{} Answer:}} The question does not provide enough information to answer directly. However, from the given paragraphs, we can infer that the retailer headquartered in Poole, Dorset, United Kingdom is Lush Ltd. It is not specified in the text whether Lush Ltd. has a store in ABQ Uptown. To answer the question, we would need additional information about the retailers in ABQ Uptown. 

Sub-questions:
1. Which retailer is headquartered in Poole, Dorset, United Kingdom? (Answer: Lush Ltd.)
2. Does Lush Ltd. have a store in ABQ Uptown? (Answer: Not specified in the text, additional information needed.) \coloredcheckmark

{\color{red}Next, we present an example where our proposed \model{} made an incorrect prediction, in contrast to the correct prediction made by 0-CoT.}

\underline{\textbf{Question:}} Paragraph 0 1965 Mexican Grand Prix: The 1965 Mexican Grand Prix was a Formula One motor race held at Ciudad Deportiva Magdalena Mixhuca in Mexico City on October 24, 1965.It was race 10 of 10 in both the 1965 World Championship of Drivers and the 1965 International Cup for Formula One Manufacturers.The race was won by Richie Ginther, who took his first victory and the first for the Honda team, after leading for the entire race.The Brabham-Climax of Dan Gurney finished the race second and the Lotus-Climax of Mike Spence completed the podium.

Paragraph 1 2006 FIA Formula One World Championship: The 2006 FIA Formula One World Championship was the 60th season of FIA Formula One motor racing.It featured the 2006 FIA Formula One World Championship which began on 12 March and ended on 22 October after eighteen races.The Drivers' Championship was won by Fernando Alonso of Renault F1 for the second year in a row, with Alonso becoming the youngest ever double world champion at the time.Then-retiring multiple world champion Michael Schumacher of Scuderia Ferrari finished runner-up, 13 points behind.The Constructors' Championship was won by Mild Seven Renault F1 Team, which defeated Scuderia Ferrari Marlboro by five points.

Paragraph 2 1963 United States Grand Prix: The 1963 United States Grand Prix was a Formula One motor race held on October 6, 1963, at the Watkins Glen Grand Prix Race Course in Watkins Glen, New York.It was race 8 of 10 in both the 1963 World Championship of Drivers and the 1963 International Cup for Formula One Manufacturers.The 110-lap race was won by BRM driver Graham Hill after he started from pole position.His teammate Richie Ginther finished second and Lotus driver Jim Clark came in third.

Paragraph 3 Formula One World Champions: A Formula One World Champion is a racing driver or automobile constructor which has been designated such a title by the governing body of Formula One - the FIA.Every Formula One World Champion since the inaugural World Drivers' Championship in 1950 and the inaugural World Constructors' Championship in 1958 has been awarded the title by accumulating the required points during the course of the F1 season of that particular year, by participating in relevant Grands Prix.

Paragraph 4 2000 FIA Formula One World Championship: The 2000 FIA Formula One World Championship was the 54th season of FIA Formula One motor racing.It featured the 2000 FIA Formula One World Championship which commenced on 12 March 2000, and ended on 22 October after seventeen races.Michael Schumacher became Ferrari's first World Drivers' Champion for 21 years having clinched the Drivers' title at the penultimate race of the season.Ferrari successfully defended its Constructors' title.This season marked the first for future world champion Jenson Button.

Paragraph 5 1963 German Grand Prix: The 1963 German Grand Prix was a Formula One motor race held at Nürburgring on August 4, 1963.It was race 6 of 10 in both the 1963 World Championship of Drivers and the 1963 International Cup for Formula One Manufacturers. \hl{The 15-lap race was won by Ferrari driver John Surtees after he started from second position.Jim Clark finished second for the Lotus team and BRM driver Richie Ginther came in third.}

Paragraph 6 1964 Austrian Grand Prix: The 1964 Austrian Grand Prix was a Formula One World Championship motor race held at Zeltweg Airfield on August 23, 1964.It was race 7 of 10 in both the 1964 World Championship of Drivers and the 1964 International Cup for Formula One Manufacturers.The 105-lap race was won by Ferrari driver Lorenzo Bandini after he started from seventh position.Richie Ginther finished second for the BRM team and Brabham driver Bob Anderson came in third.This was the debut World Championship race of the future world champion Jochen Rindt.

Paragraph 7 1962 Italian Grand Prix: The 1962 Italian Grand Prix was a Formula One motor race held at Monza on 16 September 1962.It was race 7 of 9 in both the 1962 World Championship of Drivers and the 1962 International Cup for Formula One Manufacturers.The 86-lap race was won by BRM driver Graham Hill after he started from second position.His teammate Richie Ginther finished second and Cooper driver Bruce McLaren came in third.

Paragraph 8 Richie Ginther: Paul Richard "Richie" Ginther (Granada Hills, California, August 5, 1930 – September 20, 1989 in France) was a racecar driver from the United States.During a varied career, the 1965 Mexican Grand Prix saw Ginther take Honda's first Grand Prix victory, a victory which would also prove to be Ginther's only win in Formula One.Ginther competed in 54 World Championship Formula One Grand Prix races and numerous other non-Championship F1 events.

Paragraph 9 John Surtees: John Surtees, {'1': ", '2': ", '3': ", '4': "} (11 February 1934 – 10 March 2017) was an English Grand Prix motorcycle road racer and Formula One driver.He was a four-time 500cc motorcycle World Champion – winning that title in 1956, 1958, 1959 and 1960 – the Formula One World Champion in 1964, and remains the only person to have won World Championships on both two and four wheels.He founded the Surtees Racing Organisation team that competed as a constructor in Formula One, Formula 2 and Formula 5000 from 1970 to 1978.He was also the ambassador of the Racing Steps Foundation.

Which Formula One World Champion had a teammate named Richie Ginther?

\underline{\textbf{0-CoT Answer:}} Graham Hill had a teammate named Richie Ginther. \coloredcheckmark

\underline{\textbf{\model{} Answer:}} To answer the question, we need to first identify the teams that Richie Ginther was a part of. Then, we need to find out who were his teammates in these teams and whether any of them were Formula One World Champions. 

Sub-questions:

1. Which teams did Richie Ginther race for?

2. Who were Richie Ginther's teammates on these teams?

3. Which of these teammates were Formula One World Champions?

Let's tackle these sub-questions one by one.

1. From Paragraph 2, we know that Richie Ginther was a driver for the BRM team. From Paragraph 5, we learn that he was also a driver for the Ferrari team.

2. \hl{From Paragraph 2, we know that Graham Hill was a teammate of Richie Ginther in the BRM team.} \coloredcheckmark \hl{From Paragraph 5, we know that John Surtees was a teammate of Richie Ginther in the Ferrari team.}\coloredcross

3. From Paragraph 3, we know that a Formula One World Champion is a racing driver who has been awarded the title by accumulating the required points during the course of the F1 season. From Paragraph 9, we know that John Surtees was a Formula One World Champion in 1964.

\hl{By connecting the answers from these sub-questions, we can conclude that the Formula One World Champion who had Richie Ginther as a teammate was John Surtees.}\coloredcross

\fbox{\parbox{0.9\linewidth}{
\textbf{Analysis:} As pointed out specifically in Paragraph 5 from the original context, it's noted that both John Surtees and Richie Ginther participated in the German Grand Prix. However, John Surtees was a member of the Ferrari team, while Richie Ginther raced for BRM. This information indicates that John Surtees and Richie Ginther were not teammates.}}

\subsubsection{Graph Sorting}
{\color{blue}First, we showcase an example accurately predicted by our proposed \model{}, in contrast to its incorrect prediction by 0-CoT.}

\underline{\textbf{Question:}} The following paragraphs each describe a set of three objects arranged in a fixed order. The statements are logically consistent within each paragraph. On a branch, there are three birds: a falcon, an owl, and a raven. The raven is to the left of the owl. The falcon is the leftmost. Options:

(A) The falcon is the second from the left

(B) The owl is the second from the left

(C) The raven is the second from the left

\underline{\textbf{0-CoT Answer:}} (B) The owl is the second from the left \coloredcross

\underline{\textbf{\model{} Answer:}} The knowledge graph from the story would look like this:

1. Falcon - Position: Leftmost

2. Owl - Position: Middle

3. Raven - Position: Rightmost

From the story, we know that the falcon is the leftmost, the raven is to the left of the owl, which means the owl is to the right of the raven, and since the falcon is the leftmost, the raven must be in the middle and the owl must be the rightmost.

So, evaluating the options:

(A) The falcon is the second from the left - This is incorrect as the falcon is the leftmost.

(B) The owl is the second from the left - This is incorrect as the owl is the rightmost.

(C) The raven is the second from the left - This is correct as the raven is in the middle, which makes it second from the left.

Therefore, the correct statement is (C) The raven is the second from the left. \coloredcheckmark

{\color{red}Next, we present an example where our proposed \model{} made an incorrect prediction, in contrast to the correct prediction made by 0-CoT.}

\underline{\textbf{Question:}} The following paragraphs each describe a set of three objects arranged in a fixed order. The statements are logically consistent within each paragraph. A fruit stand sells three fruits: peaches, pears, and mangoes. The mangoes are less expensive than the pears. The peaches are the cheapest.
Options:

(A) The peaches are the second-most expensive

(B) The pears are the second-most expensive

(C) The mangoes are the second-most expensive

\underline{\textbf{0-CoT Answer:}} (C) The mangoes are the second-most expensive \coloredcheckmark

\underline{\textbf{\model{} Answer:}} (B) The pears are the second-most expensive

Explanation: From the story, we can extract the following facts to create a knowledge graph:

1. The fruit stand sells peaches, pears, and mangoes.

2. The mangoes are less expensive than the pears.

3. The peaches are the cheapest.

\hl{From these facts, we can infer the order of the fruits from cheapest to most expensive: peaches, mangoes, pears. \coloredcheckmark Therefore, the pears are the second-most expensive. } \coloredcross

\fbox{\parbox{0.9\linewidth}{
\textbf{Analysis:} As we highlight, while the inference process of our proposed model is accurate, it makes mistakes during the conclusion-drawing phase. This pattern has been widely observed in our experiments.}}

\subsubsection{Graph Query}
{\color{blue}First, we showcase an example accurately predicted by our proposed \model{}, in contrast to its incorrect prediction by 0-CoT.}

\underline{\textbf{Question:}} Here is a table where the first line is a header and each subsequent line is a penguin:  name, age, height (cm), weight (kg) Louis, 7, 50, 11 Bernard, 5, 80, 13 Vincent, 9, 60, 11 Gwen, 8, 70, 15  For example: the age of Louis is 7, the weight of Gwen is 15 kg, the height of Bernard is 80 cm.  And here is a similar table, but listing giraffes:

name, age, height (cm), weight (kg)

Jody, 5, 430, 620

Gladys, 10, 420, 590

Marian, 2, 310, 410

Donna, 9, 440, 650

What is the name of the last penguin sorted by alphabetic order? Options:

(A) Louis

(B) Bernard

(C) Vincent

(D) Gwen

(E) James

\underline{\textbf{0-CoT Answer:}} (D) Gwen \coloredcross

\underline{\textbf{\model{} Answer:}} (C) Vincent

Explanation: The knowledge graph from the story would include the following facts about the penguins:

- Louis is 7 years old, 50 cm tall, and weighs 11 kg.

- Bernard is 5 years old, 80 cm tall, and weighs 13 kg.

- Vincent is 9 years old, 60 cm tall, and weighs 11 kg.

- Gwen is 8 years old, 70 cm tall, and weighs 15 kg.

When we sort these penguins by their names in alphabetical order, we get Bernard, Gwen, Louis, and Vincent. Therefore, the last penguin sorted by alphabetic order is Vincent. \coloredcheckmark

{\color{red}Next, we present an example where our proposed \model{} made an incorrect prediction, in contrast to the correct prediction made by 0-CoT.}

\underline{\textbf{Question:}} Here is a table where the first line is a header and each subsequent line is a penguin:  name, age, height (cm), weight (kg) Louis, 7, 50, 11 Bernard, 5, 80, 13 Vincent, 9, 60, 11 Gwen, 8, 70, 15  For example: the age of Louis is 7, the weight of Gwen is 15 kg, the height of Bernard is 80 cm.  How many penguins are less than 8 years old and weight more than 12 kg? Options:

(A) 1

(B) 2

(C) 3

(D) 4

(E) 5

\underline{\textbf{0-CoT Answer:}} (A) 1 \coloredcheckmark

\underline{\textbf{\model{} Answer:}} (B) 2

Explanation: 
First, let's create a knowledge graph from the given data:

1. Louis: Age - 7, Height - 50 cm, Weight - 11 kg

2. Bernard: Age - 5, Height - 80 cm, Weight - 13 kg

3. Vincent: Age - 9, Height - 60 cm, Weight - 11 kg

4. Gwen: Age - 8, Height - 70 cm, Weight - 15 kg

Now, let's answer the question: "How many penguins are less than 8 years old and weight more than 12 kg?"

From the knowledge graph, we can see that the penguins that meet these criteria are:

\hl{1. Bernard: Age - 5 (less than 8), Weight - 13 kg (more than 12 kg)

2. Gwen: Age - 8 (not less than 8), Weight - 15 kg (more than 12 kg)

3. Louis: Age - 7 (less than 8), Weight - 11 kg (not more than 12 kg)

4. Vincent: Age - 9 (not less than 8), Weight - 11 kg (not more than 12 kg)} \coloredcheckmark

\hl{So, the only penguins that meet both criteria are Bernard and Gwen.} Therefore, the answer is 2.\coloredcross

\fbox{\parbox{0.9\linewidth}{
\textbf{Analysis:} As we highlight, while the inference process of our proposed model is accurate, it makes mistakes during the conclusion-drawing phase. This pattern has been widely observed in our experiments.
}}

\subsubsection{Logical Inference}
{\color{blue}First, we showcase an example accurately predicted by our proposed \model{}, in contrast to its incorrect prediction by 0-CoT.}

\underline{\textbf{Question:}} sent1: if something is in the sunlight then that something will absorb solar energy sent2: water is in the solid state , called ice sent3: the sun is the source of solar energy called sunlight sent4: water is in the liquid state, called water sent5: an ice cube is placed in the sunlight sent6: the ice cube will melt and change from solid to liquid by absorbing the solar energy sent7: absorbing sunlight causes objects to heat A student placed an ice cube on a plate in the sun. Ten minutes later, only water was on the plate. Which process caused the ice cube to change to water?

\underline{\textbf{0-CoT Answer:}} The process that caused the ice cube to change to water is the absorption of solar energy, which caused the ice cube to heat and melt. \coloredcross

\underline{\textbf{\model{} Answer:}} To answer this question, let's first create a knowledge graph from the given story:

1. (something, in, sunlight) -> (something, absorb, solar energy)

2. (water, is in, solid state) -> (water, called, ice)

3. (sun, is source of, solar energy) -> (sun, called, sunlight)

4. (water, is in, liquid state) -> (water, called, water)

5. (ice cube, placed in, sunlight)

6. (ice cube, will melt and change from, solid to liquid) -> (ice cube, by absorbing, solar energy)

7. (absorbing sunlight, causes, objects to heat)

Now, let's transform the question into a triple: (ice cube, change to, ?). We need to find the process that caused the ice cube to change to water.

From the knowledge graph, we can see that when 'something' (in this case, the ice cube) is in the sunlight, it will absorb solar energy (Fact 1). The ice cube is in the sunlight (Fact 5), so it will absorb solar energy. Absorbing solar energy causes the ice cube to melt and change from solid to liquid (Fact 6). This change from solid to liquid is called melting. 

So, by following the logical path in the knowledge graph and using deductive reasoning, we can conclude that the process that caused the ice cube to change to water is 'melting'. \coloredcheckmark

{\color{red}Next, we present an example where our proposed \model{} made an incorrect prediction, in contrast to the correct prediction made by 0-CoT.}

\underline{\textbf{Question:}} sent1: sunlight is a kind of solar radiation sent2: ultraviolet light causes sunburn sent3: sunlight contains ultraviolet light Which form of solar radiation causes sunburn?

\underline{\textbf{0-CoT Answer:}} Ultraviolet light \coloredcheckmark

\underline{\textbf{\model{} Answer:}} To answer the question, let's first create a knowledge graph from the given sentences:

1. (sunlight, is a kind of, solar radiation)

2. (ultraviolet light, causes, sunburn)

3. (sunlight, contains, ultraviolet light)

Now, let's transform the question into a triple (s, r, ?): (Which form of solar radiation, causes, sunburn?)

The subject entity 's' here is 'Which form of solar radiation'. The relation 'r' is 'causes'. We need to find the tail entity '?' that is 'sunburn'.

Starting with the subject entity, we look for a statement related to 'solar radiation'. We find it in the first sentence: (sunlight, is a kind of, solar radiation). So, 'sunlight' is a form of 'solar radiation'.

Next, we look for a statement related to 'sunlight'. We find it in the third sentence: (sunlight, contains, ultraviolet light). So, 'sunlight' contains 'ultraviolet light'.

Finally, we look for a statement related to 'ultraviolet light'. We find it in the second sentence: (ultraviolet light, causes, sunburn). \hl{So, 'ultraviolet light' causes 'sunburn'.} \coloredcheckmark

By combining these conclusions, we can deduce that 'sunlight', which is a form of 'solar radiation' and contains 'ultraviolet light', causes 'sunburn'.

\hl{In conclusion, the form of solar radiation that causes sunburn is 'sunlight'.} \coloredcross

\fbox{\parbox{0.9\linewidth}{
\textbf{Analysis:} As we highlight, while the inference process of our proposed model is accurate, it makes mistakes during the conclusion-drawing phase. This pattern has been widely observed in our experiments.
}}

\subsection{Prompts}\label{app:Prompts}
We provide the prompts that we used to query the LLMs for all six tasks in Tables~\ref{tab:relational reasoning prompt} to \ref{tab:logical inference prompt}. We do not use the system message field for any model. To facilitate recognition, each stage of the prompt is color-coded: {\color{olive}olive for the first stage}, {\color{teal}teal for the second}, and {\color{violet}violet for the third}.

\subsection{Raw Results}\label{app:raw results}
We show the numeric results for all six tasks in Tables~\ref{tab:Results over CLUTRR} to \ref{tab:Results over Entailment Bank}.

\begin{table*}
\centering
\caption{Prompts for relation prediction.} 
\label{tab:relational reasoning prompt}
\begin{tabularx}{\textwidth}{c|X}
\toprule[1.5pt] 
Mode & Prompt \\
\hline
0-CoT & Based on the story, through inductive reasoning think step by step to answer the question.\\
\hline
\model{} & {\color{olive}First, create a knowledge graph by extracting facts from each sentence in the given input story.} {\color{teal} Once this is done, I will pose a question. This question can be transformed into a triple (s, ?, o), where your primary task is to determine the missing relation ('?') that links the subject entity ('s') to the object entity ('o'). To begin, focus on the subject entity in this triple and choose the most relevant facts to expand from it. Step by step, progress towards the object entity, ensuring that each selected fact contributes to creating a link between the subject and object entities.} {\color{violet}Finally, utilize the established connection between the subject and object entities to answer the question.} \\
\bottomrule[1.5pt] 
\end{tabularx}
\end{table*}

\begin{table*}
\centering
\caption{Prompts for entity prediction over dynamic KG.} 
\label{tab:dynamic KG prompt}
\begin{tabularx}{\textwidth}{c|X}
\toprule[1.5pt] 
Mode & Prompt \\
\hline
0-CoT & Given the input, think step by step to answer the question using the option number. \\
\hline
\model{} & {\color{olive}First, create a knowledge graph by extracting facts from each sentence in the given input story. The graph should evolve as the story progresses.} {\color{teal}  I will present several statements. Your primary task is to determine the correctness of these statements by converting them into triples (s, r, o).} {\color{violet} Assess each statement's validity against the knowledge graph as it stands at the story's conclusion.} \\
\\
\bottomrule[1.5pt] 
\end{tabularx}
\end{table*}

\begin{table*}
\centering
\caption{Prompts for complex entity prediction.} 
\label{tab:complex entity prediction prompt}
\begin{tabularx}{\textwidth}{c|X}
\toprule[1.5pt] 
Mode & Prompt \\
\hline
0-CoT & Based on the story, think step by step to answer the question. \\
\hline
\model{} & {\color{olive}First, create a knowledge graph by extracting facts from each sentence in the given input story.} {\color{teal} Once this is done, I will pose a complex question requiring multi-step reasoning. Decompose the question into simpler sub-questions focusing on identifying crucial entities, their relationships, and specific details. Tackle these sub-questions sequentially, referencing the knowledge graph for information.} {\color{violet} Connect the answers from these sub-questions step by step, until arrive at a final answer to the initial complex question.}
\\
\bottomrule[1.5pt] 
\end{tabularx}
\end{table*}

\begin{table*}
\centering
\caption{Prompts for graph sorting.} 
\label{tab:graph sorting prompt}
\begin{tabularx}{\textwidth}{c|X}
\toprule[1.5pt] 
Mode & Prompt \\
\hline
0-CoT & Given the input, think step by step to answer the question using the option number. \\
\hline
\model{} & {\color{olive}First, create a knowledge graph by extracting facts from each sentence in the given input story.} {\color{teal} Once this is done, I will provide several statements. Your primary task is to determine the correctness of these statements.} {\color{violet} To assess the validity of a statement, sort the objects in the graph and evaluate the correctness of each statement.}
\\
\bottomrule[1.5pt] 
\end{tabularx}
\end{table*}

\begin{table*}
\centering
\caption{Prompts for graph query.} 
\label{tab:graph query prompt}
\begin{tabularx}{\textwidth}{c|X}
\toprule[1.5pt] 
Mode & Prompt \\
\hline
0-CoT & Given the input, think step by step to answer the question using the option number. \\
\hline
\model{} & {\color{olive}First, create a knowledge graph by extracting facts from each sentence in the given input story. The graph may evolve as the story progresses.} {\color{teal}Once this is done, I will pose a question. This question will require you to identify objects that meet specific criteria.} {\color{violet} Utilize the final state of the knowledge graph, as it exists at the end of the story, to provide the answer to the question.}
\\
\bottomrule[1.5pt] 
\end{tabularx}
\end{table*}

\begin{table*}
\centering
\caption{Prompts for logical inference.} 
\label{tab:logical inference prompt}
\begin{tabularx}{\textwidth}{c|X}
\toprule[1.5pt] 
Mode & Prompt \\
\hline
0-CoT & Based on the story, through deductive reasoning think step by step to answer the question. \\
\hline
\model{} & {\color{olive}First, create a knowledge graph by extracting facts from each sentence in the given input story.} {\color{teal} Once this is done, I will pose a question. This question can be transformed into a triple (s, r, ?), where your primary task is to determine the missing tail entity ('?') that connects the subject entity ('s') through the relation (’r’). Start by concentrating on the subject entity in this triple and follow a logical path within the knowledge graph. Progress step by step from the statement related to the subject, using a forward chaining process. At each step, combine the conclusions with the facts in the knowledge graph to deduce new conclusions.} {\color{violet} The final conclusion along this logical path will serve as the answer to the question.}
\\
\bottomrule[1.5pt] 
\end{tabularx}
\end{table*}

\begin{table*}
\centering
\caption{Results over CLUTRR.} 
\label{tab:Results over CLUTRR}
\begin{tabular}{c|c|c|c|c|c|c|c|c|c}
\toprule[1.5pt] 
\multicolumn{2}{c|}{Mode} & 3 hop & 4 hop & 5 hop & 6 hop & 7 hop & 8 hop & 9 hop & 10 hop \\
\Xhline{1pt}
\multirow{3}{*}{GPT 3.5} & w/o 0-CoT & 0.162 & 0.097 & 0.062 & 0.033 & 0.023 & 0.015 & 0.05
& 0.041 \\
\cline{2-10}
& w/ 0-CoT & 0.295 & 0.143 & 0.041 & 0.022 & 0.023 & 0.015 & 0.02 & 0.031 \\
\cline{2-10}
& \model{} & \textbf{0.562} & \textbf{0.422} & \textbf{0.329} & \textbf{0.267} & \textbf{0.233} & \textbf{0.195} & \textbf{0.26} & \textbf{0.258} \\
\Xhline{1pt}
\multirow{3}{*}{GPT 4} & w/o 0-CoT & 0.543 & 0.331 & 0.247 & 0.244 & 0.116 & 0.120 & 0.09 & 0.124 \\
\cline{2-10}
& w/ 0-CoT & 0.543 & 0.325 & 0.247 & 0.3 & 0.116 & 0.135 & 0.11 & 0.093 \\
\cline{2-10}
& \model{}& \textbf{0.695} & \textbf{0.604} & \textbf{0.568} & \textbf{0.5} & \textbf{0.434} & \textbf{0.406} & \textbf{0.39} & \textbf{0.299} \\
\bottomrule[1.5pt] 
\end{tabular}
\end{table*}

\begin{table*}
\centering
\caption{Results over BIG-bench-hard.} 
\label{tab:Results over BIG-bench-hard}
\begin{threeparttable} 
\begin{tabular}{cc|c|c|c|c|c|c|c|c}
\toprule[1.5pt] 
\multicolumn{2}{c|}{\multirow{2}{*}{Mode}} &\multicolumn{3}{c|}{Dynamic KG} & \multicolumn{3}{c|}{Graph Sorting} & \multicolumn{2}{c}{Graph Query} \\
\cline{3-10}
& & \makecell{tracking \\ shuffled \\objects \\ three \\ objects} & \makecell{tracking \\ shuffled \\ objects \\ five \\ objects}
& \makecell{tracking \\ shuffled \\ objects \\ seven \\ objects} & 
\makecell{logical \\ deduction \\ three \\ objects} & \makecell{logical \\ deduction \\ five \\ objects}
& \makecell{logical \\ deduction \\ seven \\ objects} & \makecell{reasoning \\ about \\ colored \\ objects} & \makecell{penguins \\ in \\ a \\ table} \\
\Xhline{1pt}
\multirow{3}{*}{GPT 3.5} & w/o 0-CoT & 0.32 & 0.16 & 0.128 & 0.572 & 0.4 & 0.416 & 0.56 & 0.692 \\
\cline{2-10}
& w/ 0-CoT &0.384 &0.28 & 0.232 &0.66 &0.54 &0.632& 0.684 &0.815 \\
\cline{2-10}
&\model{}  &\textbf{0.816} &\textbf{0.84} &\textbf{0.836} & \textbf{0.84} &\textbf{0.904} &\textbf{0.856} 
& \textbf{0.808} &\textbf{0.877} \\
\Xhline{1pt}
\multirow{3}{*}{GPT 4} & w/o 0-CoT &0.484 &0.34 &0.304 &0.952 &0.676 &0.652 & 0.88 &0.760 \\
\cline{2-10}
& w/ 0-CoT & 0.44 &0.324 &0.28 &0.932 &0.664 &0.636 & 0.86 &0.753 \\
\cline{2-10}
& \model{} &\textbf{0.936} &\textbf{0.924} &\textbf{0.916} & \textbf{0.968} &\textbf{0.86} &\textbf{0.824} 
& \textbf{0.94} &\textbf{0.884} \\
\bottomrule[1.5pt] 
\end{tabular}
\end{threeparttable}
\end{table*}

\begin{table*}
\centering
\caption{Results over HotpotQA.} 
\label{tab:Results over HotpotQA}
\begin{threeparttable} 
\begin{tabular}{c|c|c|c|c|c|c}
\toprule[1.5pt] 
\multirow{2}{*}{Mode} &\multicolumn{3}{c|}{GPT 3.5} &\multicolumn{3}{c}{GPT 4} \\
\cline{2-7}
& w/o 0-CoT  & \makecell{w/ 0-CoT} & \makecell{\model{}} & w/o 0-CoT & \makecell{w/ 0-CoT} & \makecell{\model{}}\\
\hline
ACC & 0.42 & 0.407 & \textbf{0.44} &0.427 & 0.407 & \textbf{0.527}
\\
\bottomrule[1.5pt] 
\end{tabular}
\end{threeparttable}
\end{table*}

\begin{table*}
\centering
\caption{Results over Entailment Bank.} 
\label{tab:Results over Entailment Bank}
\begin{tabular}{c|c|c|c|c|c|c|c}
\toprule[1.5pt] 
\multicolumn{2}{c|}{Mode} & 1 hop & 2 hop & 3 hop & 4 hop & 5 hop & 6 hop \\
\Xhline{1pt}
\multirow{3}{*}{GPT 3.5} & w/o 0-CoT & \textbf{0.674} &0.489 &0.444 &0.377 &0.227 &\textbf{0.429}\\
\cline{2-8}
& w/ 0-CoT & 0.628 &0.5& \textbf{0.472} & 0.396 & 0.273 & \textbf{0.429} \\
\cline{2-8}
&\model{} & 0.640 &\textbf{0.533} & 0.431 & \textbf{0.453} & \textbf{0.364} & \textbf{0.429} \\
\Xhline{1pt}
\multirow{3}{*}{GPT 4} & w/o 0-CoT & 0.663 &\textbf{0.611} &0.403 &0.453 &0.227 &\textbf{0.429} \\
\cline{2-8}
& w/ 0-CoT & 0.663 &0.589 &0.403 &\textbf{0.491} &0.227 &0.357 \\
\cline{2-8}
& \model{}& \textbf{0.709} &0.544 &\textbf{0.472} &0.453 &\textbf{0.318} &\textbf{0.429}\\
\bottomrule[1.5pt] 
\end{tabular}
\end{table*}